\documentclass[preprint,12pt]{elsarticle}

\usepackage{algorithm}
\usepackage[noend]{algorithmic}
\usepackage{booktabs}
\usepackage{float}
\usepackage{amssymb}
\usepackage{amsmath,mathtools}
\usepackage{multirow}
\usepackage[pagebackref,breaklinks,colorlinks]{hyperref}
\usepackage[normalem]{ulem}
\useunder{\uline}{\ul}{}
\usepackage[dvipsnames]{xcolor}

\journal{Computers \& Graphics}

\begin{document}

\begin{frontmatter}



\title{Domain-adaptive Zero-Shot Image Enhancement via Locality-Constrained Diffusion Guidance 
}

\tnotetext[note1]{%
Accepted manuscript. The final version is published in \textit{Computers \& Graphics} and available at
\url{https://doi.org/10.1016/j.cag.2026.104607} 
 \\ \textcopyright\ 2026. This manuscript version is made available under the CC BY-NC-ND 4.0 license.
}

\author{Theresa~Neubauer\fnref{label1}}

\author{Dimitrios~Lenis\fnref{label1}}
\author{Astrid~Berg\fnref{label1}}
\author{Maria~Wimmer\fnref{label1}}

\author{Gaia~Romana~De~Paolis\fnref{label1}}
\author{Philip~Matthias~Winter\fnref{label1}}
\author{David~Major\fnref{label1}}
\author{Johannes~Novotny\fnref{label1}}
\author{Ariharasudhan~Muthusami\fnref{label1}}

\author{Katja~Bühler\fnref{label1}}

\affiliation[label1]{organization={VRVis GmbH, Vienna}, country={Austria}}


%\author{} %% Author name

%\affiliation{organization={},%Department and Organization

%% Abstract

\begin{abstract}

Denoising Diffusion Probabilistic Models have shown remarkable performance in unconditional image generation. In order to generate images with desired semantics, recent works have restricted the solution space by using guidance constraints in the diffusion sampling process. 

However, for image enhancement across different domains, these methods struggle to balance two main requirements: looking realistic in the target domain (photorealistic images) and preserving relevant features of the source domain, e.g., low-quality renderings or art paintings. Here, small local changes can alter the fidelity of the image completely, while large changes in other regions might be insignificant. 

We introduce \textit{LocDiff}, a locality-constrained guidance method for image enhancement, which serves as a zero-shot extension to pre-trained diffusion models, ensuring the preservation of critical features during domain adaptation. 
In this way, we retain important local features, while allowing less critical regions to remain unconstrained and not interfere with the guidance process for relevant regions.
We evaluate our method on two different domain-shift tasks:
For art-to-photo translation, we apply the method in a fully zero-shot setting, preserving facial identity from paintings while generating photorealistic details.
For enhancing low-quality fetal ultrasound renderings, we demonstrate zero-shot inference with auxiliary prior alignment. Here, the objective is to artificially add high-resolution characteristics and produce photorealistic ultrasound renderings, a target domain for which no ground truth distribution exists.
Our experimental results demonstrate that LocDiff achieves favorable realism-faithfulness trade-offs compared to state-of-the-art methods, enabling controllable cross-domain enhancement.

\end{abstract}

%% disable for arxiv version
%\includegraphics[trim=0 0 0 0, clip,width=1\linewidth]%{images/graphical_abstract2.pdf}

%% Keywords

\end{frontmatter}



\begin{figure*}[h!]
\includegraphics[width=1\textwidth]{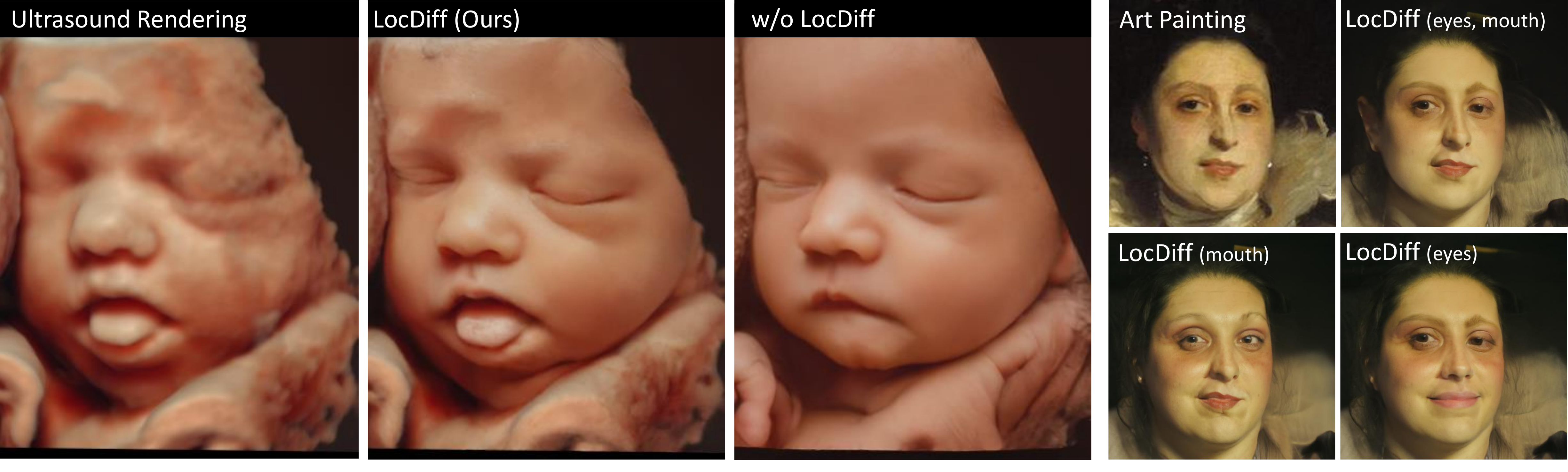}
			\caption{Our proposed locality-constrained guidance method named LocDiff controls the generation process of diffusion models with region-adaptive conditions, allowing the user fine-grained control of image enhancement, even under domain shifts. \textbf{Left:}  We enhance pre-trained diffusion models with LocDiff to preserve facial characteristics of low-quality fetal ultrasound renderings. 
            \textbf{Right:}  Locality-constrained guidance to enhance face parts of art paintings with different strengths (constrained face parts in brackets).
			}
   \label{fig:intro_figure}
\end{figure*}

\section{Introduction}
\label{sec:intro}

\noindent Deep learning techniques have revolutionized image generation, opening up new ways of creative expression to the general public without the need for technical expertise (Stable Diffusion~\cite{rombachHighResolutionImageSynthesis2022}, DALL-E 2~\cite{rameshHierarchicalTextConditionalImage2022}). 
While these tools often \textit{generate} content from text prompts describing the desired outcome, there is also a strong need for \textit{enhancement of existing images}, made possible through techniques from inpainting and editing to super-resolution~\cite{sahariaPaletteImageImageDiffusion2022, mengSDEDITGUIDEDIMAGE2022}. Current Generative Adversarial Networks (GAN) and diffusion-based methods are producing images that appear increasingly realistic \textit{(realism)} \cite{rombachHighResolutionImageSynthesis2022}. However, for successful image enhancement of a source image, it is essential that resulting images preserve the distinctive features of the source \textit{(faithfulness)}~\cite{mengSDEDITGUIDEDIMAGE2022}, like structural and color details. Existing approaches for photo-realistic image enhancement often struggle to meet these requirements simultaneously~\cite{wangRealWorldBlindFace2021, yueDifFaceBlindFace2024a}, with methods trading off between generating realistic details and preserving source identity.
GAN-based methods such as conditional GANs and GAN inversion methods offer limited control to preserve source-domain specific semantics \cite{isolaImageImageTranslationConditional2017,  shenInterpretingLatentSpace2020a}. Additionally, they lack robustness when exposed to domain shifts, necessitating retraining on the new domain~\cite{zhuUnpairedImageImageTranslation2017, choiStarGANUnifiedGenerative2018, karrasStyleBasedGeneratorArchitecture2019a}.

Recent advances in diffusion-based zero-shot image generation leverage the generative prior of pre-trained diffusion models with additional guidance, enabling finer control of the generation process without the need for retraining~\cite{wangZeroShotImageRestoration2022, feiGenerativeDiffusionPrior2023}. 
However, these methods primarily prove effective for inputs that align with the domain of the trained diffusion prior, such as denoising blurred face images originating from the high-resolution portrait domain~\cite{wangZeroShotImageRestoration2022}, while neglecting input images exhibiting significant domain shifts.

For image enhancement, a low-quality input image is provided, with the objective to enhance it with high-quality visual features. In contrast to the regular diffusion process, starting from random noise, the low-quality image is introduced at an intermediate sampling step with noise levels corresponding to this step. This process makes models robust to various types of degradation and to some extent acts as domain adaptation~\cite{choiILVRConditioningMethod2021}. However, images that exhibit significant distribution shifts from the diffusion prior, which is in most cases the photo-realistic image domain, typically require an increased number of diffusion steps to achieve high-resolution features, as higher noise levels are necessary to align the input image with the learned distribution. These higher noise levels can compromise faithfulness, leading to weaker correspondence of fine details and diminished preservation of crucial structures, such as faces~\cite{feiGenerativeDiffusionPrior2023}. Therefore, it is essential to implement guidance mechanisms that preserve the underlying structure as accurately as possible while producing a realistic natural image. 

We propose a diffusion guidance mechanism for domain-adaptive zero-shot image enhancement on unseen target domains.
Following \cite{kodirovUnsupervisedDomainAdaptation2015, changpinyoSynthesizedClassifiersZeroShot2016}, we define zero-shot learning as inference on target domains that remain entirely unobserved during training, crucially without any access to domain-specific samples. We use \textit{domain-adaptive }to describe adjusting a model's implicit prior under this zero-shot constraint. While the proposed guidance operates without any retraining of the diffusion model, we show that under strong semantic mismatch (fetal ultrasound renderings), performance can be substantially improved through an optional prior-alignment step using auxiliary real-image datasets. Importantly, this alignment step does not involve any target-domain data, and inference on the target domain remains zero-shot.

In this paper, we propose \textit{LocDiff}, a method to balance realism and faithfulness requirements of image generation methods, particularly under domain shifts.
Unlike existing zero-shot image enhancement techniques that rely on guidance applied uniformly across the entire image, we address domain shift tasks by introducing locality-constrained guidance. This approach enables region-specific flexibility, allowing for diverse conditioning methods and varying guidance strengths to be applied across different image areas.
In this way, we can ensure that important local features are preserved, while allowing less critical regions, such as the background, to remain unconstrained and not interfere with the guidance process for other relevant regions.

\noindent \textbf{Contributions}:

\begin{itemize}
	\item We introduce \textit{LocDiff}, a novel locality-constrained guidance method. Through utilizing flexible conditioning regions and an adaptive sampling schedule for individual image regions, our method acts as a zero-shot extension to the reverse diffusion process, able to selectively preserve critical details during domain adaptation from the input to the diffusion prior domain.
 
 	\item  Our ablation results indicate that we can precisely target specific image regions by independently modulating the guidance for this region in the reversed diffusion process. 
	
	\item We demonstrate the effectiveness of \textit{LocDiff} across distinct domains: 1)~art-to-photo and 2)~rendering-to-photo. We evaluate \textit{LocDiff} using image restoration, where the aim is to preserve the facial identity and relevant features of the source domain, while at the same time adding high-frequency details of the photorealistic domain. 
    Our experimental results demonstrate that \textit{LocDiff} performs competitively with state-of-the-art methods on art-to-photo and rendering-to-photo tasks, achieving effective realism-faithfulness trade-offs on both art painting and ultrasound rendering datasets.

    \item \textit{LocDiff} supports both fully zero-shot scenarios (art-to-photo without fine-tuning) and zero-shot inference with auxiliary prior alignment (ultrasound task with semantic space alignment), ensuring flexibility across domains without requiring target-domain training data.

\end{itemize}

\section{Related Work}
\label{sec:related_work}

\noindent \textbf{Guidance Strategies for Diffusion Models.}
Denoising Diffusion Probabilistic Models \cite{sohlDeepUnsupervisedLearning2015} are known for generating impressive photorealistic images from random noise.
In the forward diffusion process, Gaussian noise is iteratively added to an input image. In the reverse sampling process, the diffusion model learns to reverse the forward process and thereby iteratively denoises the data, aiming to reconstruct the original image.
Previous studies indicate that diffusion models are inherently adept at learning the data manifold~\cite{pidstrigachScoreBasedGenerativeModels2022, stanczukDiffusionModelsEncode2022}. For their guidance strategies, these works utilize the assumption that high-dimensional data points (such as images) form an (assumed linear) low-dimensional manifold \cite{bengioRepresentationLearningReview2013}.
Typically, this results in a two-step approach~\cite{chungDiffusionPosteriorSampling2022,  chungImprovingDiffusionModels2022, heManifoldPreservingGuided2024}: a denoising step, which performs an orthogonal projection onto the data manifold, followed by a guidance step that advances tangentially along the current manifold. 
Recent works have focused on developing guidance strategies for pre-trained diffusion models to adapt them to different downstream tasks, using various guidance concepts such as text, styles, sketches, or facial attributes~\cite{zhangSketchGuidedTextImageGeneration2024, tumanyanPlugPlayDiffusionFeatures2023, guFilterGuidedDiffusionControllable2024d, yuFreeDoMTrainingFreeEnergyGuided2023a, heManifoldPreservingGuided2024}.

To selectively guide specific regions of an image, some methods employ binary masks, such as in inpainting~\cite{lugmayrRePaintInpaintingUsing2022} or text-to-image tasks~\cite{bar2023multidiffusion}. 
However, these approaches offer limited control over the enhancement process.
For example, MultiDiffusion~\cite{bar2023multidiffusion} uses binary masks to guide image generation with text prompts, starting from random noise, thus can't be used for enhancing existing images. RePaint~\cite{lugmayrRePaintInpaintingUsing2022} applies binary masks during the reverse diffusion process to inpaint the selected regions with new content, while leaving the rest of the image unchanged. 
Critically, these binary masking approaches operate in an all-or-nothing manner, regions are either fully constrained or fully unconstrained, and lack mechanisms to preserve structural consistency within guided regions. Our approach differs by enhancing the entire image with adaptive, region-specific guidance strength that enables fine-grained control over the realism-faithfulness trade-off, preserving both structure and visual coherence across all regions.\\

\noindent \textbf{Diffusion Prior for Image Enhancement with Domain Shifts.}
Existing methods for image enhancement~\cite{yueDifFaceBlindFace2024a, zhuDenoisingDiffusionModels2023, cui2024taming} or image editing \cite{sahariaPaletteImageImageDiffusion2022, mengSDEDITGUIDEDIMAGE2022} show that low quality input images (from a different domain) can be mapped to the high-quality image domain of a pre-trained unconditioned diffusion model by adding various conditions to the diffusion sampling process. 
Various approaches have been proposed: 
(1)~Classifier guidance adjusts the intermediate steps in the sampling process, by incorporating the gradient of the log-likelihood derived from an auxiliary classifier~\cite{dhariwalDiffusionModelsBeat2021}, which has demonstrated effectiveness in tasks such as image enhancement~\cite{yangPGDiffGuidingDiffusion2023, yuFreeDoMTrainingFreeEnergyGuided2023a}. Nevertheless, classifier guidance lacks generalizability to unseen classes and introduces considerable training overhead, limiting its suitability for domain adaptation tasks.
(2)~Other methods~\cite{chungDiffusionPosteriorSampling2022, choiILVRConditioningMethod2021, feiGenerativeDiffusionPrior2023, wangZeroShotImageRestoration2022, lin2024diffbir} use the low-quality input image to guide pre-trained diffusion models by adjusting the intermediate steps in the sampling process, aligning the intermediate output more closely to the guide.
Recent approaches have also explored frequency-domain decomposition, combining low-frequency and high-frequency priors through wavelet transforms~\cite{Shang2024multidomain}.
(3)~An alternative approach to constrain the solution space of the diffusion model involves starting the reverse sampling process at an intermediate step, rather than beginning with random noise. Specifically, the process starts with the low-quality image and adds noise corresponding to the intermediate step~\cite{mengSDEDITGUIDEDIMAGE2022, yueDifFaceBlindFace2024a}.
Consequently, images exhibiting mild degradation will require fewer diffusion steps, whereas images with stronger degradation or significant domain shifts will necessitate more steps \cite{fabianAdaptDiffuseSampleAdaptive2024}. 

These guidance methods apply their strategies uniformly across the entire image. However, for domain adaptation, not all regions of the image carry the same importance. Therefore, we propose adapting constraints for each region based on its relevance to ensure faithfulness in the enhanced image. This approach preserves even small local features while allowing less critical areas to remain unconstrained, ensuring better adaptation to the diffusion prior's domain.\\

\noindent \textbf{Image Enhancement for Faces.}
A variety of face enhancement techniques employ face-specific constraints to direct the generation process, thereby facilitating the generation of enhanced facial features, such as generative priors~\cite{wangRealWorldBlindFace2021, menonPULSESelfSupervisedPhoto2020, lin2024diffbir}, facial landmarks or parsing maps \cite{chenProgressiveSemanticAwareStyle2021, chenFSRNetEndEndLearning2018}, or codebooks with mappings from low quality and high quality images as reference priors~\cite{zhouRobustBlindFace2022, wangRestoreFormerRealWorldBlind2023a, guVQFRBlindFace2022}, or image editing with text-prompts~\cite{brooks2023instructpix2pix, mokady2023nulltext}.
However, GAN-based restoration models often perform poorly when the input image or degradation type differs from what they were trained on. The work by Kuai et al.~\cite{kuai2025blindface} enhances the model's robustness by training directly on real-world degraded images. However, if the target domain deviates from photorealistic imagery, retraining is necessary to maintain the enhancement quality.
In contrast, recent studies \cite{yueDifFaceBlindFace2024a, miaoWaveFaceAuthenticFace2024, fabianAdaptDiffuseSampleAdaptive2024, choiILVRConditioningMethod2021, lin2024diffbir} leveraging diffusion priors demonstrate higher robustness to small domain shifts, yielding visually realistic results.
Degraded or stylistically different images, which exhibit significant distribution shifts from the photorealistic face domain, typically require higher noise levels (i.e., more diffusion steps) to produce satisfactory results within the target distribution \cite{fabianAdaptDiffuseSampleAdaptive2024}. However, increased noise can result in weaker anatomical correspondence.
Given the ability of humans to discern slight differences in facial features, even small differences can result in a different face~\cite{maurerManyFacesConfigural2002}.
To preserve facial identity, it is crucial to maintain fine-grained structural information in key regions, such as eyes, nose, and mouth.
Blind face restoration methods, as discussed in \cite{qiuDiffBFRBootstrappingDiffusion2023a, yangPGDiffGuidingDiffusion2023, zhaoAuthenticFaceRestoration2023}, employ various techniques to preserve facial identities. However, these methods have been exclusively designed for the photorealistic domain. In contrast, our proposed method is designed to handle domain shifts in the context of image enhancement.

\section{Method}
\label{sec:method}
\label{sec:background}
\noindent We address the challenge of enhancing low-quality images from a shifted domain (e.g., artistic paintings, ultrasound renderings) by leveraging pre-trained diffusion models trained on high-quality photorealistic images. 
Our method, LocDiff (Locality-Constrained Diffusion), introduces region-specific guidance during the reverse diffusion process. Instead of applying uniform constraints across the entire image, which either over-preserve the input (losing realism) or over-enhance it (losing identity), we apply constraints with different guidance strengths to different facial regions. 
This is especially relevant for translation between domains, where different regions of an image require different levels of enhancement: critical facial features should be preserved to maintain faithfulness, while less important areas can be freely enhanced to achieve photorealism. Through disjoint binary masks, our method makes it possible to apply strong preservation constraints to regions critical for identity like eyes, while background regions receive weak constraints. 

We now formalize this approach and provide the technical details.

\noindent \textbf{Problem Setup.}
We assume possession of a small set of low-quality images $\mathcal{Y} \coloneqq \{ y \}_{i=1}^N$, that we aim to enhance. As is common for real-world scenarios, no corresponding high-quality ground-truth is available (cf. Fig.~\ref{fig:intro_figure}). 
We propose to leverage a pre-trained diffusion model trained on a high-quality target distribution $p_\mathrm{data}(x)$, assuming semantic continuity despite a clear distribution shift between $ \mathcal{Y} \sim p(y)$ and $p_\mathrm{data}$. For example, ultrasound renderings of fetuses and photorealistic baby photographs show similar subjects but distinct visual characteristics like skin texture.
Our goal is to enhance the quality of a sample $y \sim p(y)$, by mapping it towards $p_\mathrm{data}$. This transformations aims to preserve the original characteristics of $y$ (the \emph{low frequency} features) ensuring \emph{faithfulness}, while enriching it with high-frequency characteristic of $p_\mathrm{data}$, hence improving \emph{realism}. For this, we propose \emph{locality-constrained guidance} to bridge the shift between $p(y)$ and $p_\mathrm{data}$, overcoming the unconstrained transform of the diffusion prior (cf. Section~\ref{sec:comparison_sota} and Figs.~\ref{fig:bf_tstart} and~\ref{fig:bf_baselines}). \\

\noindent \textbf{Background on Diffusion Models}. For a given dataset $\mathcal{I} = \{x_i\}_{i=1}^{N'} \sim p_ \mathrm{data}(x) $, a diffusion model can capture the implicit prior of the underlying data distribution $p_ \mathrm{data}$ by aligning with the gradient of the log density (known as the score function $ \nabla_{x} \mathrm{log}\,p_t(x_t) $) \cite{chungDiffusionPosteriorSampling2022}. Conceptually, these models consist of a forward process (the diffusion) of $T$ steps, and its reverse.
Let $y \in \mathcal{Y}$ be a sample of interest. For $x_i \in \mathcal{I}$ we abbreviate $x := x_i$, such that its iterations $x_{i,t}$ simplify to $x_t$, $t\in [0, T]$, similarly for $y$.
In the forward process, $\{p_t\}_{0 \leqslant t \leqslant T}$, the model progressively adds noise following the transition kernel $ p(x_t \mid x_0) = \mathcal{N}(x_0 \alpha_t ,\sigma_t^2 \mathbb{I})$ (with $\alpha_t$ and $\sigma_t$ being scalar functions characterizing the noise schedule). Starting from data distribution $p_0(x) = p_\mathrm{data} $, this process gradually erodes the underlying data structure, eventually approaching the standard Gaussian distribution $p_T(x) \approx \mathcal{N}(0;\mathbb{I})$ as $T \to \infty$. 
The reverse process learns to denoise by fitting a neural network $s_\theta(x_t, t)$ that approximates the score function, resulting in samples sharing the characteristics of the data-distribution $ p_\mathrm{data}$. 
In SOTA denoising diffusion models, this is characterized by a stochastic differential equation, which can serve as a generative model once the score function $ \nabla_{x} \mathrm{log}\,p_t(x_t) $ is estimated tractably~\cite{chungDiffusionPosteriorSampling2022}. This again is done by fitting a neural network $s_\theta(x_t,t)$ via stochastic regression, leveraging $\mathbb{E}_{t, x, p_{t,x_0}} \left[ \| p(x_t \mid x_0  - s_\theta(x_t, t)) \|_2 \right]$ \cite{heManifoldPreservingGuided2024}. A classic result building upon Tweedie's formula~\cite{song2023pseudoinverse} shows that an estimate for the clean data $x_{0|t}$ can be derived through $x_{0|t}~=~(\sqrt{\alpha_t})^{-1} (x_t - \sqrt{1 - \alpha_t} \, s_\theta(x_t,t)) $. \\

\subsection{Locality-Constrained Guidance}
\begin{figure*}[h!]
	\centering
	\includegraphics[width=1\linewidth]{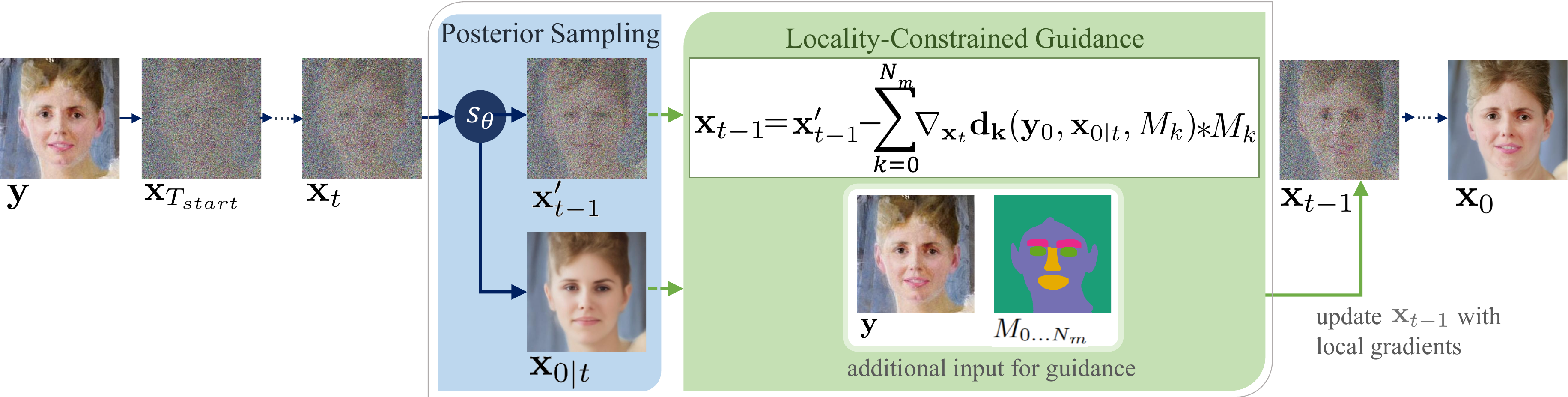}

	\caption{Visualization of the locality-constrained guidance step. 
		For each conditioning region $M_k$ (user-defined or model generated binary masks), the distance function $d_k$ measures the distance between the input $y$ and the predicted $x_{0|t}$ restricted to $M_k$. % at time step $t$. 
        For each region $M_k$, the gradient of $d_k$ with respect to $x_t$ is computed independently and updates the current $x_{t-1}$ exclusively within the  region $M_k$.}
	\label{fig:overview}
\end{figure*}

In this section, we introduce \textit{LocDiff}, a local manifold guidance method for controlling the reverse sampling process of an unconditional pre-trained diffusion model.
Our method extends the reverse diffusion sampling process with a locality-constrained guidance step comprising two parts:
(1) sampling the posterior of the pre-trained diffusion model (ensuring \emph{realism}), and (2) refining it with a localized gradient to constrain the solution space to local manifolds that align with the input image (maintaining \emph{faithfulness}). Figure~\ref{fig:overview} illustrates our method. \\

\noindent \textbf{Posterior Sampling for Realism.}
To generate samples conditioned on input $y$, we rewrite the conditional score  $\nabla_{x} \mathrm{log}\,p_t(x_t|y) =  \nabla_{x_t} \mathrm{log}\, p(x_t) + \nabla_{x_t} \mathrm{log}\, p(y| x_t)$  (Bayes).
The first term is directly approximated by our pre-trained diffusion model $s_\theta(x_t,t)$ by sampling from the posterior  $p_\mathrm{data}$, through $ x_{t-1} = ( -\frac{1}{2} \alpha_t x_{t-1} - \alpha_t ( \nabla_{x_t} \log p_t(x_t) + \nabla_{x_t} \log p_t(y \mid x_t)) ) + \sqrt{\alpha_t} \sigma_t $ \cite{chungDiffusionPosteriorSampling2022}, where $\alpha$ and $\sigma$ are components of the noise schedule (cf. above), while the second term (the likelihood) guides the sample toward compatibility with $y$.

The noisy likelihood gradient serves as an optimization step through gradient descent to minimize the guidance loss in the neighborhood of the denoised sample $x_t$, pointing towards solutions compatible with $y$ \cite{chungDiffusionPosteriorSampling2022}.
Following diffusion posterior sampling, we approximate the noisy likelihood using the clean data estimate  $ \nabla_{x_t} \log p(y \mid x_t) \simeq \nabla_{x_t} \log p(y \mid x_{0|t}) $ with a controllable upper bound \cite{chungDiffusionPosteriorSampling2022}. In combination, this provides a per-timestep guidance given by $\nabla_{x_t} \log p_t(x_t \mid y) \approx s_{\theta}^*(x_t, t) - \zeta_t \nabla_{x_t}  \log \, p(y \mid x_{0|t}) $, where $\zeta_t$ scales the step size.

While this guides sampling towards realism within the diffusion model's distribution, uniform image-level conditioning allows the learned prior to dominate fine-grained details, potentially distorting critical structures and diminishing semantic correspondence.

As full image conditioning fails in this regard (cf. Fig.~\ref{fig:comparison_locality-constrained_vs_entire_image}), we propose to combine manifold constraining with \emph{locality-constrained guidance}. \\

\noindent \textbf{Locality-Constrained Guidance for Faithfulness.}

 The local guidance, \textit{LocDiff}, consists of three components: (i) a set of attribution maps $\mathcal{M}$, (ii)~a differentiable distance measure $d$, and (iii) a sampling scheme for guidance timesteps $t_g$. 
An attribution map $\mathcal{M} \coloneqq \{M_k\}_{k=0}^{N_m}$ identifies a set of regions to constrain toward either the source or target domain. By selecting which timesteps $t_g$ apply local guidance and choosing the distance function $d$, we control constraint strength from strong preservation (faithfulness) to weak constraints (realism).

Inspired by the product of experts formulations~\cite{kongDiffusionModelsConstrained2024} where the target density is modeled as a product of individual components, we interpret our desired distribution as favoring configurations with high probability across all regions. Since $ \nabla_x \log \pi(x) = \sum_e \nabla_x \log \pi_e(x) $, which is compatible with standard sampling strategies \cite{kongDiffusionModelsConstrained2024}, we refine the constrained denoised estimate $x_{t-1}$ through the sum of region-specific guidance steps: \\ $x_{t-1} - \sum_{k=0}^{N_m}   \zeta_t \nabla_{x_t}  d(y, x_{0|t}, M_k) * M_k $, where $\zeta_t$ is a time dependent dampening factor. 

\noindent \textbf{Adaptive Timestep Selection.}
The sampling scheme $t_g$ is based on the observation that at some timestep $\tau$, the noisy input $y_\tau$ aligns with noisy samples from $ p_\mathrm{data} $: $x_\tau \approx y_\tau$.
Depending on the observed gap between the data manifolds of the input data and the training data for the diffusion model, we carefully design our sampling scheme to only apply local guidance at selected timesteps $t_g$. 
Consequently, input images that are similar to the training distribution require fewer diffusion steps (less noise) to bridge the gap between the two data manifolds, whereas images with significant domain shifts necessitate more steps~\cite{fabianAdaptDiffuseSampleAdaptive2024}. As the forward process advances, high-frequency details are progressively eliminated, causing neighboring samples from distinct manifolds to converge in appearance. Adapting semantic features (low-frequency features) to the input $y$ should therefore be performed at later timesteps, specifically $t \geq \tau$. 
If the gap between the two data manifolds is too large (high $\tau$), fine-tuning the diffusion prior on a small subset of data that more closely aligns with the input domain can significantly improve image enhancement. Algorithm~\ref{alg:locality-constrained} describes our steps, where masks $M$ in this algorithm are binary, disjoint, and zero on non-relevant regions.

\begin{algorithm}[]
\small
	\caption{\textbf{Locality-Constrained Sampling Process}}
	\begin{algorithmic}[1]
		\renewcommand{\algorithmicrequire}{\textbf{Require:}}
		
		\REQUIRE 	original image $y$, score network $s_{\theta}(\cdot,t)$ \\ 
		\phantom{aaaaa} distance function $d(y,\cdot,M)$ \\
		\phantom{aaaaa} pre-defined parameters $\beta_t$, $\bar{\alpha}_t$, $\zeta_t$, $\tilde\sigma_t$ \\

		\phantom{aaaaa} list of conditioning regions $L=[(M, d, t_{g}),...]$ \\
        \phantom{aaaaa} \textbf{Constraint:} masks $M$ are binary and pairwise disjoint \\
		\phantom{aaaaa} starting timestep $T_{start}$ for $1 \leq T_{start} < T$

		\STATE Sample $x_{T_{start}}$ from $q(x_{T_{start}}|y)$ 
	
		\FOR {$t = T_{start}$ to $1$}
		\STATE $\epsilon_t \sim\ \mathcal{N}(0,I)$
	
		\STATE $x_{0|t}=\frac{1}{\sqrt{\bar{\alpha}_t}}(x_t+(1-\bar{\alpha}_t)s_{\theta}(x_t, t))$
		
		\STATE{$x_{t-1}' = \frac{\sqrt{\bar\alpha_t}(1-\bar\alpha_{t-1})}{1 - \bar\alpha_t}x_t + \frac{\sqrt{\bar\alpha_{t-1}}\beta_t}{1 - \bar\alpha_t}x_{0|t} +  {\tilde\sigma_t \epsilon_t}$}

		\FOR {$(M,d,t_{g})$ in $L$}
		\IF {$t$ in $t_{g}$}
		\STATE $g=\nabla_{x_{t}} d(y, x_{0|t}, M)*M$
	
		\STATE $x_{t-1}=x_{t-1}'-\zeta_t g*M$ 
		\ENDIF
		\ENDFOR
		\ENDFOR
		\STATE \textbf{return} $x_{0}$

	\end{algorithmic}
	\label{alg:locality-constrained}

\end{algorithm}

\section{Experiments}
\label{sec:experiments}

This study evaluates the impact of our locality-constrained guidance on specific image regions. We demonstrate its effectiveness in art-to-photo and rendering-to-photo translation for face images, aiming to convert low-quality (art/ultrasound) images into photorealistic ones while preserving facial identity. We focus on faces since it is a challenging task and due to the availability of precise evaluation metrics to measure facial preservation.

\subsection{Diffusion Prior for Domain-Adaptive Zero-Shot Tasks}

For artistic face portraits, we directly apply a diffusion model pre-trained on high-quality face images (FFHQ dataset~\cite{karrasStyleBasedGeneratorArchitecture2019a}), without any fine-tuning. This constitutes a fully zero-shot setting under strong domain shift.

For rendered fetal face enhancement, we observe that the FFHQ prior contains adult-specific modes that dominate the conditional generation and negatively impact the resulting baby face semantics. We therefore perform an auxiliary prior-alignment step using a small collected dataset (100 samples) of real high-resolution portrait images of babies, cf.~\ref{sec:appendix_datasets} for dataset details.
We fine-tune $s_{\theta}$ for 20,000 epochs following the training strategy of~\cite{yueDifFaceBlindFace2024a}.
Importantly, this fine-tuning does not introduce target-domain information, as the envisioned target domain (enhanced ultrasound images with clearer facial details) does not exist in the real world and is never observed during training.  In this case, the target domain is defined implicitly through a conditioning process and desired output constraints at inference time. In such settings, training on the target domain is infeasible, and inference is necessarily zero-shot with respect to the target. Due to the prior alignment, we label the task as “zero-shot inference with auxiliary prior alignment". 

We utilize the framework for conditional diffusion models 
from Dhariwal et al.~\cite{dhariwalDiffusionModelsBeat2021}. 
To accelerate the diffusion process, we employ the DDIM sampling strategy~\cite{songDenosingDiffusionImplict2021}, reducing the number of diffusion steps from 1000 to 250. 
Additionally, we leverage the pre-trained diffusion model from DifFace~\cite{yueDifFaceBlindFace2024a} as score estimator $s_{\theta}$,  trained on the FFHQ dataset~\cite{karrasStyleBasedGeneratorArchitecture2019a} for image denoising tasks, and use the linear noise schedule from~\cite{yueDifFaceBlindFace2024a} for $\beta$.

\subsection{Datasets, Implementation Details, Metrics}

\noindent \textbf{Datasets for Evaluation.} 
\textit{MetFaces}~\cite{karrasTrainingGenerativeAdversarial2020} comprises 1,366 portrait images of human faces extracted from various artworks, including paintings and sculptures. 

\textit{WikiArt-Faces} consists of images of human faces extracted from paintings available on \href{WikiArt.org}{WikiArt.org}. We curated all face portraits within the realism category, resulting in a dataset of 1,373 portrait images, cf. \ref{sec:appendix_datasets}.  

The \textit{fetal ultrasound rendering} dataset comprises 437 2D images, projected from 3D-rendered ultrasound volumes, showing fetuses in the second and third trimester. The dataset was acquired by our medical partner in accordance with ethical standards, and we have obtained consent to use it for research purposes in this study. \\ 

\noindent  \textbf{Conditioning Regions.}
\label{experiments_regions}
Conditioning regions can be derived from suitable segmentation algorithms or manually specified by the user.
For the art-to-photo task, we define five conditioning regions: 
(1) eyes, (2) eye brows, (3) mouth and nose, (4)~skin, and (5)~background (remaining image region). 
For the rendering-to-photo task, we define five conditioning regions:
1) eyes, (2) mouth, (3) nose, (4)~skin, and (5)~background. 
In this study, the attribution maps $M_{1...5}$ are generated using a face parsing model~\cite{zheng2022farl} (trained on the LaPa dataset~\cite{liu2020new}) that predicts binary masks for individual face parts. Each mask, excluding the background and skin, is extended by 20 pixels without overlap. However, the eye mask used for the rendering-to-photo task is extended by 40 pixels.
The ablation study presented in Tab.~\ref{tab:ablation} evaluates individual conditioning regions to examine their contributions to the guidance mechanism in the art-to-photo task.
For the comparison to SOTA, we use all five conditioning regions (multi-region) for guidance. \\

\noindent  \textbf{Distance Function.} For the image enhancement task, we design the distance function $\mathbf{d}$ similar to Chung et al.~\cite{chungDiffusionPosteriorSampling2022}, but restrict the image region for the corresponding conditioning region with an attribution map $M$:
$\mathbf{d}(y, x_{0|n}, M)\coloneq\|y*M-\mathcal{A}(x_{0|n}*M) \|_2^2$.
The operator $\mathcal{A}$ is defined as a Gaussian blur kernel with a size of $5 \times 5$ and standard deviation of 4. 
Using a low-pass filter is common for guidance methods with  low-quality reference images to focus on coarser structures for distance calculation \cite{yuFreeDoMTrainingFreeEnergyGuided2023a, choiILVRConditioningMethod2021, chungDiffusionPosteriorSampling2022}. 
The distance function was selected empirically, as justified in Tab.~\ref{tab:ablation_hyperparameter}. \\

\noindent  \textbf{Guidance Strength.}
To reduce the guidance strength, there are two strategies: (1) Set a lower value for the hyperparameter $\zeta_t$ to decrease the influence of the guidance gradient, or (2) restrict the guidance to certain steps $t_g$ in the reverse sampling process. 
We use $\zeta_t$ = 1.0 for all experiments and reduce the number of guidance steps to minimize the costly gradient calculations. 
We observe that image quality improves when guidance for certain time steps is skipped. Our observation aligns with the findings of Yu et al. \cite{yuFreeDoMTrainingFreeEnergyGuided2023a}
and Gu et al. \cite{guVQFRBlindFace2022}, who noted that guidance in the last sampling steps (refinement stage) harms the development of high-frequency details. 
We define the time step schedule $t_g$ for guidance of one conditioning region as:
$t_g= [{ i \in \mathbb{N} \mid t_{min} \leq i \leq T_{start} \text{ and } i \equiv 0 \ (\text{mod} \ m) } ]$. Details regarding the art-to-photo and the rendering-to-photo task are described in \ref{sec:appendix-art2photo} and \ref{sec:appendix-rendering2photo}.\\

\noindent  \textbf{Evaluation Metrics.} 
To evaluate faithfulness for faces, we use the identity preservation metrics Identity Score (IDS)~\cite{dengArcFaceAdditiveAngular2019} and Landmark Distance (LMD)~\cite{yueDifFaceBlindFace2024a}.
We adapt the SSIM metric to evaluate image similarity for face-specific regions, naming it SSIM-F.
To assess realism we use the Fréchet Inception Distance (FID) \cite{heuselGANStrained2017} to measure the similarity between the feature distributions of our predicted images and the FFHQ dataset~\cite{karrasStyleBasedGeneratorArchitecture2019a}, a dataset consisting of high-quality face images. For the rendering-to-photo task, we use the collected baby portrait dataset for calculating the FID metric.
We also evaluate image similarity using PSNR and LPIPS, which are widely employed for image enhancement.
LMD was omitted for rendering-to-photo, as no reliable landmark detection model exists for fetal ultrasound data.
To select the best hyperparameters for our locality-constrained guidance method, we define a metric to quantify the balance between realism and faithfulness: $\text{RF} = (1 - \text{SSIM-F}) \cdot \text{FID}$, where FID quantifies image quality and $(1 - \text{SSIM-F})$ measures content preservation. 
See \ref{sec:appendix_metrics} for implementation details.

\noindent  \textbf{Hardware Setup.} 
All experiments were conducted on a Linux Debian server equipped using an Intel Xeon Gold 5118 CPU and an Nvidia A100 GPU (40GB VRAM).

\subsection{Ablation on Locality-Constrained Guidance} 
\label{section:ablation}
\noindent \textbf{Conditioning regions.}
To demonstrate the effectiveness of the locality-constrained guidance, we perform an ablation study on the art datasets, in which conditioning is applied to different image regions. 
The following experiments were performed: 
(\MakeUppercase{\romannumeral 1}) no locality-constrained guidance, 
(\MakeUppercase{\romannumeral 2}) joint region for mouth and nose (single condition), 
(\MakeUppercase{\romannumeral 3}) separate conditions for eye and eyebrow region (two conditions), 
(\MakeUppercase{\romannumeral 4}) full-image region (single condition) using the same $t_g$ as in  (\MakeUppercase{\romannumeral 2}), 
(\MakeUppercase{\romannumeral 5}) the proposed multi-region guidance: eye, eyebrow, mouth and nose, skin, and background regions.
Individual LMD scores are calculated for eye, mouth, and nose separately, presented in Tab.~\ref{tab:ablation}. 
As expected, face-specific LMD scores increase for unconstrained face parts.

In Tab.~\ref{tab:ablation} we observe the realism-faithfulness trade-off. The model without locality-constrained guidance achieves a high image quality score (FID), but low scores in terms of identity preservation (IDS, LMD, SSIM-F) and image similarity (LPIPS, PSNR). 
More extensive guidance enhances the model's ability to maintain the original identity and similarity of the images, albeit at the expense of lower FID.

\begin{table*}[]
\centering
\scriptsize
\begin{tabular}{llllllllll}
\multicolumn{1}{l|}{}                                                                            & \multicolumn{6}{c|}{\textit{faithfulness}}                                                                                         & \multicolumn{1}{c|}{\textit{realism}} & \multicolumn{2}{c}{\textit{traditional metrics}} \\ \hline
\multicolumn{1}{l|}{\multirow{2}{*}{\textit{Conditioning}}}                                      & LMD$\downarrow$ & LMD             & LMD               & LMD              & SSIM-F$\uparrow$ & \multicolumn{1}{l|}{IDS$\downarrow$} & \multicolumn{1}{l|}{FID$\downarrow$}  & LPIPS$\downarrow$        & PSNR$\uparrow$        \\
\multicolumn{1}{l|}{}                                                                            &                 & eye$\downarrow$ & mouth$\downarrow$ & nose$\downarrow$ &                  & \multicolumn{1}{l|}{}                & \multicolumn{1}{l|}{}                 &                          &                       \\ \hline
\multicolumn{10}{c}{WikiArt-Faces Dataset}                                                                                                                                                                                                                                                                                       \\
\multicolumn{1}{l|}{\textit{(\MakeUppercase{\romannumeral 1}) w/o guidance}} & 8.28            & 7.78            & 8.61              & 6.80             & 0.41             & \multicolumn{1}{l|}{65.98}           & \multicolumn{1}{l|}{\textbf{56.22}}   & 0.36                     & 22.33                 \\
\multicolumn{1}{l|}{\textit{(\MakeUppercase{\romannumeral 2}) mouth \& nose}}                    & 7.15            & 6.14            & 4.57              & 3.64             & 0.49             & \multicolumn{1}{l|}{55.28}           & \multicolumn{1}{l|}{65.97}            & 0.33                     & 24.08                 \\
\multicolumn{1}{l|}{\textit{(\MakeUppercase{\romannumeral 3}) eyes}}                             & 7.58            & 3.76            & 8.13              & 5.61             & 0.50             & \multicolumn{1}{l|}{57.20}           & \multicolumn{1}{l|}{64.82}            & 0.33                     & 23.68                 \\
\multicolumn{1}{l|}{\textit{(\MakeUppercase{\romannumeral 4}) full image}}                       & 4.76            & 4.39            & 4.95              & 3.75             & 0.55             & \multicolumn{1}{l|}{47.23}           & \multicolumn{1}{l|}{72.92}            & \textbf{0.30}            & \textbf{26.61}        \\
\multicolumn{1}{l|}{\textit{(\MakeUppercase{\romannumeral 5}) multi-region}}                     & \textbf{3.91}   & \textbf{3.39}   & \textbf{4.17}     & \textbf{3.10}    & \textbf{0.61}    & \multicolumn{1}{l|}{\textbf{43.52}}  & \multicolumn{1}{l|}{70.88}            & 0.32                     & 25.21                 \\ \hline
\multicolumn{10}{c}{MetFaces Dataset}                                                                                                                                                                                                                                                                                            \\
\multicolumn{1}{l|}{\textit{(\MakeUppercase{\romannumeral 1}) w/o guidance}} & 8.72            & 8.77            & 8.67              & 6.90             & 0.38             & \multicolumn{1}{l|}{68.59}           & \multicolumn{1}{l|}{\textbf{49.19}}   & 0.31                     & 21.54                 \\
\multicolumn{1}{l|}{\textit{(\MakeUppercase{\romannumeral 2}) mouth \& nose}}                    & 7.44            & 7.08            & 4.06              & 3.09             & 0.46             & \multicolumn{1}{l|}{56.58}           & \multicolumn{1}{l|}{60.09}            & 0.29                     & 23.39                 \\
\multicolumn{1}{l|}{\textit{(\MakeUppercase{\romannumeral 3}) eyes}}                             & 7.71            & 3.32            & 8.61              & 5.91             & 0.48             & \multicolumn{1}{l|}{58.00}           & \multicolumn{1}{l|}{57.73}            & 0.29                     & 23.02                 \\
\multicolumn{1}{l|}{\textit{(\MakeUppercase{\romannumeral 4}) full image}}                       & 4.21            & 3.91            & 4.39              & 3.11             & 0.53             & \multicolumn{1}{l|}{46.61}           & \multicolumn{1}{l|}{66.44}            & \textbf{0.26}            & \textbf{25.95}        \\
\multicolumn{1}{l|}{\textit{(\MakeUppercase{\romannumeral 5}) multi-region}}                     & \textbf{3.21}   & \textbf{2.72}   & \textbf{3.44}     & \textbf{2.47}    & \textbf{0.60}    & \multicolumn{1}{l|}{\textbf{40.95}}  & \multicolumn{1}{l|}{64.76}            & 0.27                     & 24.58                 \\ \hline
\end{tabular}

\caption{Ablation study on the effectiveness of locality-constrained guidance for different image regions on WikiArt-Faces and MetFaces. Our proposed approach, locality-constrained guidance for multiple regions, demonstrates superior performance in terms of identity preservation, as evidenced by the LMD and IDS metrics. Furthermore, it exhibits an  improvement in image quality (FID) compared to full image guidance. The best results are highlighted in \textbf{bold}.}
	\label{tab:ablation}
\end{table*}

\begin{figure}[h!]
	
	\centering

	\includegraphics[trim=0 0 0 0, clip,width=0.72\linewidth]{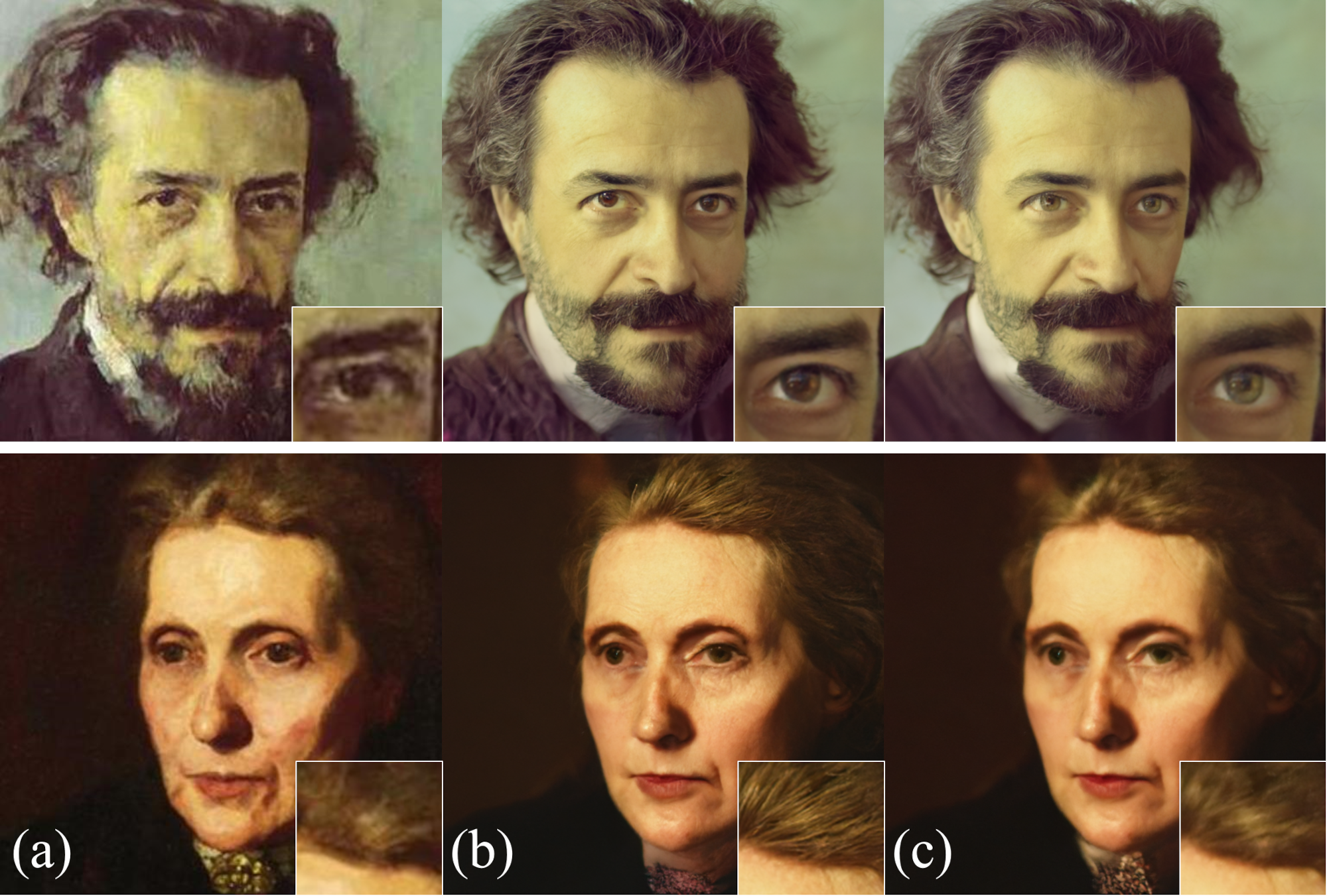}
	
	\caption{Compared to the full-image guidance \textbf{(c)}, the locality-constrained guidance \textbf{(b)} preserves the natural eye colour and lip structure of the original painting \textbf{(a)}, while the reduced guidance in the background region enhances high-frequency details for hair.}
	\label{fig:comparison_locality-constrained_vs_entire_image}
	
\end{figure}

\begin{figure}[]
	
	\centering
	
	\includegraphics[trim=0 0 0 0, clip,width=0.78\linewidth]{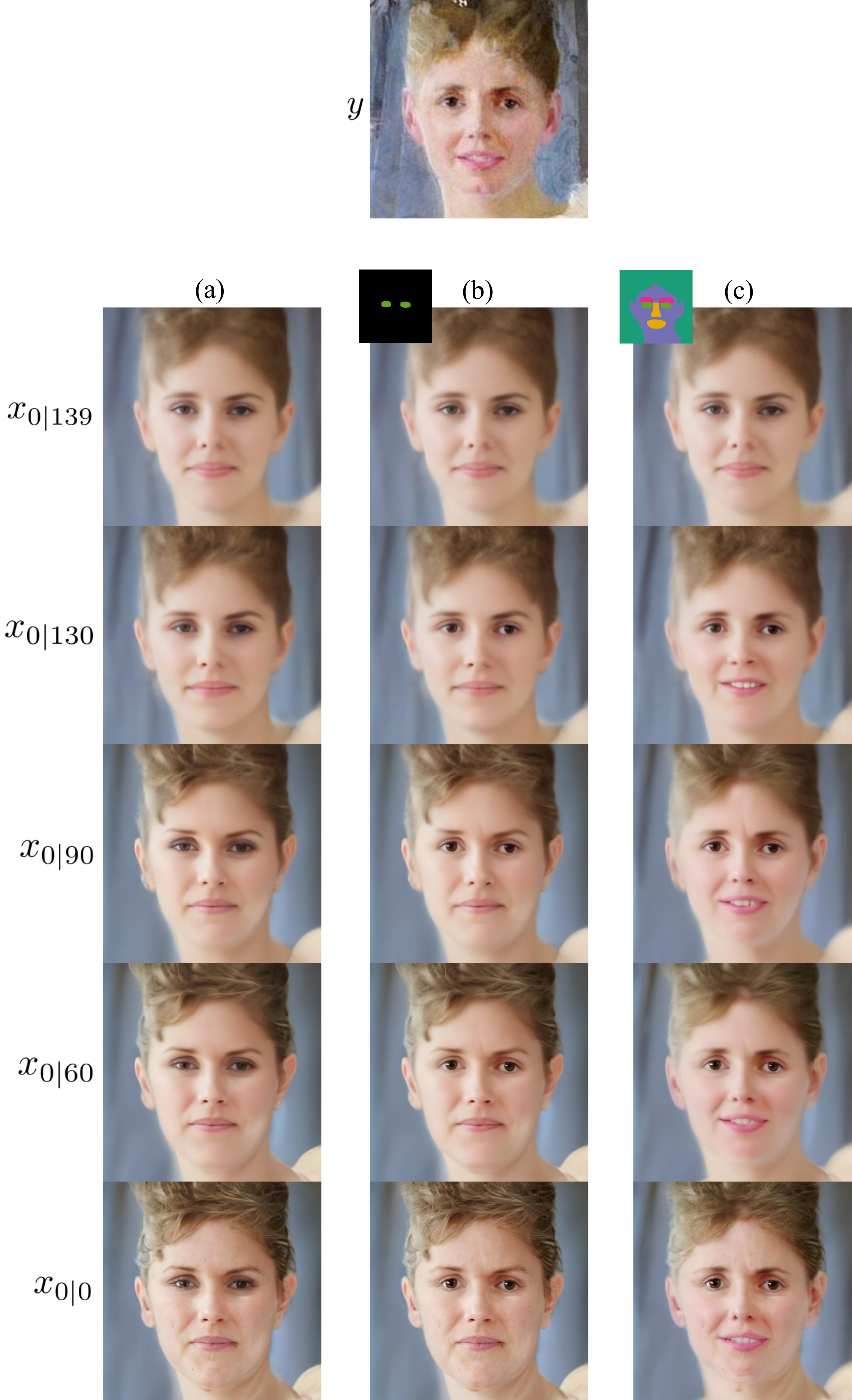}
	
	\caption{Demonstration of the locality-constrained guidance for face restoration of $y$. Images $x_{0|t}$ are generated during the reverse sampling process using no locality-constrained guidance \textbf{(a)}, locality-constrained guidance for eyes only \textbf{(b)}, locality-constrained guidance for eye, eyebrow, mouth and nose, and skin~\textbf{(c)}.
	}
	\label{fig:x0_sampling}
	
\end{figure}

\begin{figure}[h!]
	
	\centering

	\includegraphics[trim=0 0 0 0, clip,width=0.9\linewidth]             {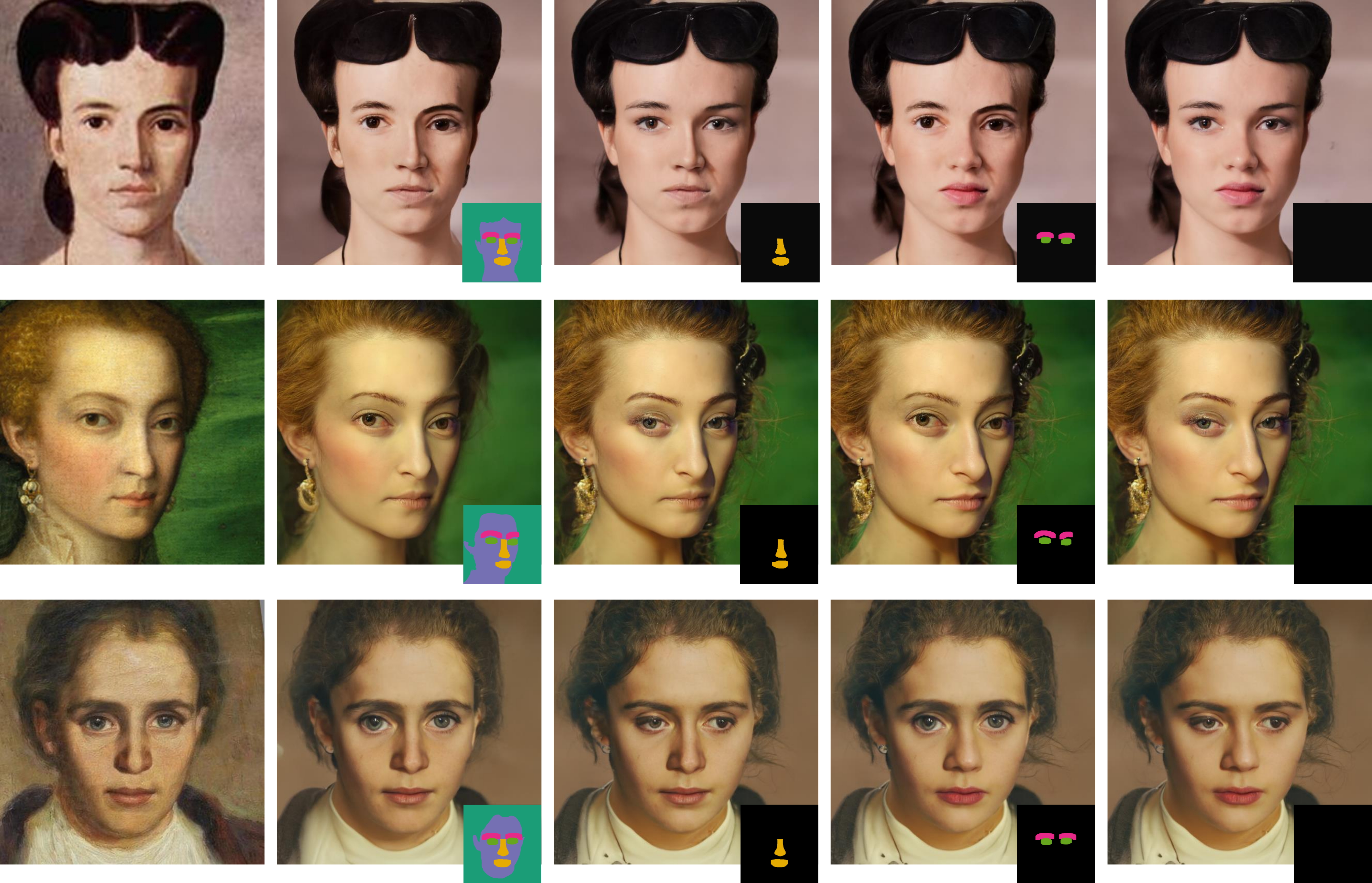}
	
	\caption{\textbf{Art-to-photo: }Qualitative results of our ablation study for single and multi-condition guidance of specific face regions using our locality-constrained guidance. Each color in the attribution map indicates a separate image region that is individually conditioned, where black regions are not guided by our locality-constrained guidance method.
	}
	\label{fig:appendix_locality_constrained_face_parts}
	
\end{figure}

More importantly, our findings indicate that our proposed multi-region guidance (\MakeUppercase{\romannumeral 5}) leads to higher identity preservation, while still maintaining enhanced image quality over the full-image guidance (\MakeUppercase{\romannumeral 4}). 
Figure~\ref{fig:comparison_locality-constrained_vs_entire_image} shows that the locality-constrained guidance excels in preserving the natural eye color of the original painting, while the reduced guidance (less guidance steps in $t_g$) in the background region enhances high-frequency details for hair.
To visualize the influence of locality-constrained guidance during the reverse sampling process, we present the intermediate results of $x_{0|t}$ in Fig.~\ref{fig:x0_sampling}. 
Guidance via conditioning regions influences only the specified areas (Fig.~\ref{fig:x0_sampling}b), allowing the remaining regions to evolve similarly to the unguided process (Fig.~\ref{fig:x0_sampling}a). Notably, the final image output exhibits no visible boundary artifacts around the conditioned regions, but instead demonstrates a smooth and coherent transition between guided and unguided areas.
Figure \ref{fig:appendix_locality_constrained_face_parts} contains further qualitative comparisons that explore how different conditioning regions impact the final result.

\textbf{Additional hyperparameters.} 
Table~\ref{tab:ablation_hyperparameter} shows the conducted ablation study on the number of diffusion steps $T_{start}$, the distance function $\mathbf{d}$, and the guidance strength $m$ for the time step guidance schedule $t_g$ (lower $m$ means stronger guidance) using the proposed multi-region conditioning from Section~\ref{section:ablation}.
For distance function $\mathbf{d}$, we tested: ($\mathbf{d_1}$) 5×5 blur, ($\mathbf{d_2}$) 9×9 blur, ($\mathbf{d_3}$) no blur, and resizing function $r$ with factors 0.25 ($\mathbf{d_4}$), 0.125 ($\mathbf{d_5}$), defined as $\|r(y* M)-r(x_{0|n}*M) \|$.

\begin{figure}[h!]
	
	\centering

	\includegraphics[trim=0 0 0 0, clip,width=1.0\linewidth]{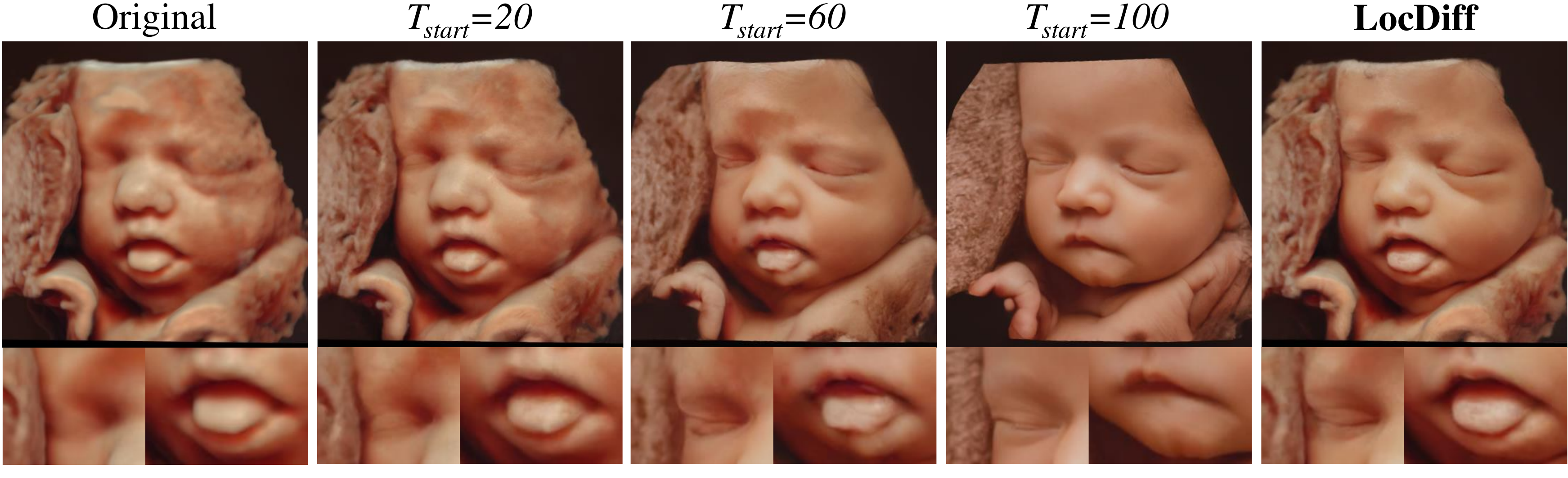}
	
	\caption{Comparison of reconstructed results without locality-constrained guidance across different starting time steps ($T_{start}$) illustrating the faithfulness-realism tradeoff. At $T_{start}=60$, eye appearance lacks realism, while at  $T_{start}=100$, mouth appearance lacks faithfulness. Our proposed locality-constrained guidance method, LocDiff, preserves facial details and maintains realism.
	}
	\label{fig:bf_tstart}
	
\end{figure}

We provide qualitative results on varying starting time steps $T_{start}$ presented in Fig.~\ref{fig:bf_tstart}. 
As $T_{start}$ increases, the generated images exhibit enhanced photorealism; however, this improvement is accompanied by a degradation in structural fidelity. For instance, at lower values of $T_{start}$, the model fails to accurately reconstruct high-resolution details for e.g. the left eye. Conversely, at 
$T_{start}=100$, the left eye appears realistic, yet other structural elements, such as the tongue, are omitted.
This phenomenon may be attributed to the scarcity of tongue-containing images in the training dataset. As $T_{start}$ increases, granting the model greater generative freedom, the diffusion prior tends to shift semantic content toward the dominant learned distribution. To mitigate this effect, our proposed multi-region guidance framework enables region-specific conditioning, allowing targeted control over individual image areas. In Fig.~\ref{fig:bf_tstart}, we apply stronger guidance to preserve structural fidelity in the mouth region, while employing weaker guidance for the eye region to promote photorealism.

\begin{table}[h!]
\scriptsize
\centering

\begin{tabular}{rll|ccc|l|l}
\multicolumn{3}{c|}{\textit{hyperparameter}}    & \multicolumn{3}{c|}{\textit{faithfulness}}           & \multicolumn{1}{c|}{\textit{realism}} & \multicolumn{1}{c}{\textit{overall}} \\ \hline
$T_{start}$      & $m$        & $\mathbf{d}$      & LMD$\downarrow$ & SSIM-F$\uparrow$ & IDS$\downarrow$ & \multicolumn{1}{c|}{FID$\downarrow$}  & \multicolumn{1}{c}{RF$\downarrow$}   \\ \hline
60           & 2            & $\mathbf{d_1}$    & 2.96 / 3.29     & 0.61 / 0.63      & 35.2 / 36.7     & 72.3 / 81.3                           & 28.2 / 30.0                          \\
100          & 2            & $\mathbf{d_1}$    & 2.96 / 3.52     & 0.61 / 0.62      & 36.5 / 39.1     & 67.1 / 72.3                           & 26.5 / 27.4                          \\
{\ul 140}    & {\ul 2}      & {\ul$\mathbf{d_1}$}  & 3.21 / 3.91     & 0.60 / 0.61      & 41.0 / 43.5     & 64.8 / 70.9                           & {\ul 26.1 / 27.4}                    \\
180          & 2            & $\mathbf{d_1}$    & 3.39 / 3.97     & 0.59 / 0.61      & 42.0 / 44.8     & 69.8 / 76.0                           & 28.4 / 29.8                          \\
220          & 2            & $\mathbf{d_1}$    & 3.29 / 3.91     & 0.60 / 0.61      & 42.5 / 44.9     & 66.1 / 74.0                           & 26.6 / 28.7                          \\ \hline
140          & 1            & $\mathbf{d_1}$    & 2.97 / 3.60     & 0.62 / 0.63      & 39.1 / 41.8     & 69.1 / 75.5                           & 26.2 / 27.6                          \\
140          & 4            & $\mathbf{d_1}$    & 3.97 / 4.37     & 0.56 / 0.57      & 45.2 / 46.6     & 62.9 / 68.8                           & 27.9 / 29.4                          \\
140          & 8            & $\mathbf{d_1}$    & 5.07 / 5.20     & 0.52 / 0.54      & 52.3 / 52.3     & 61.4 / 67.9                           & 29.6 / 30.9                          \\ \hline
140          & 2            & $\mathbf{d_2}$    & 3.57 / 3.95     & 0.56 / 0.58      & 45.1 / 44.4     & 63.1 / 70.9                           & 27.6 / 30.0                          \\
140          & 2            & $\mathbf{d_3}$    & 3.15 / 3.74     & 0.62 / 0.64      & 40.6 / 43.0     & 68.4 / 75.4                           & 26.3 / 27.5                          \\
140          & 2            & $\mathbf{d_4}$    & 5.23 / 5.28     & 0.51 / 0.53      & 54.8 / 54.1     & 56.2 / 61.9                           & 27.5 / 28.9                          \\
140          & 2            & $\mathbf{d_5}$    & 6.64 / 6.31     & 0.45 / 0.48      & 61.0 / 59.0     & 53.0 / 58.8                           & 28.9 / 30.7                          \\ \hline
140          & \multicolumn{2}{l|}{no guidance} & 8.72 / 8.28     & 0.38 / 0.41      & 68.6 / 66.0     & 49.2 / 56.2                           & 30.9 / 33.0                         
\end{tabular}

\caption{Ablation study on the number of diffusion steps $T_{start}$, guidance strength $m$, and distance function $\mathbf{d}$ for the locality-constrained guidance, evaluated on MetFaces/WikiArt datasets. Stronger guidance (lower $m$ indicates stronger guidance) improves faithfulness but harms realism. The best setting is underlined.
}
\label{tab:ablation_hyperparameter}
\end{table}

\begin{figure}[h!]
	
	\centering

	\includegraphics[trim=0 0 0 0, clip,width=0.96\linewidth]{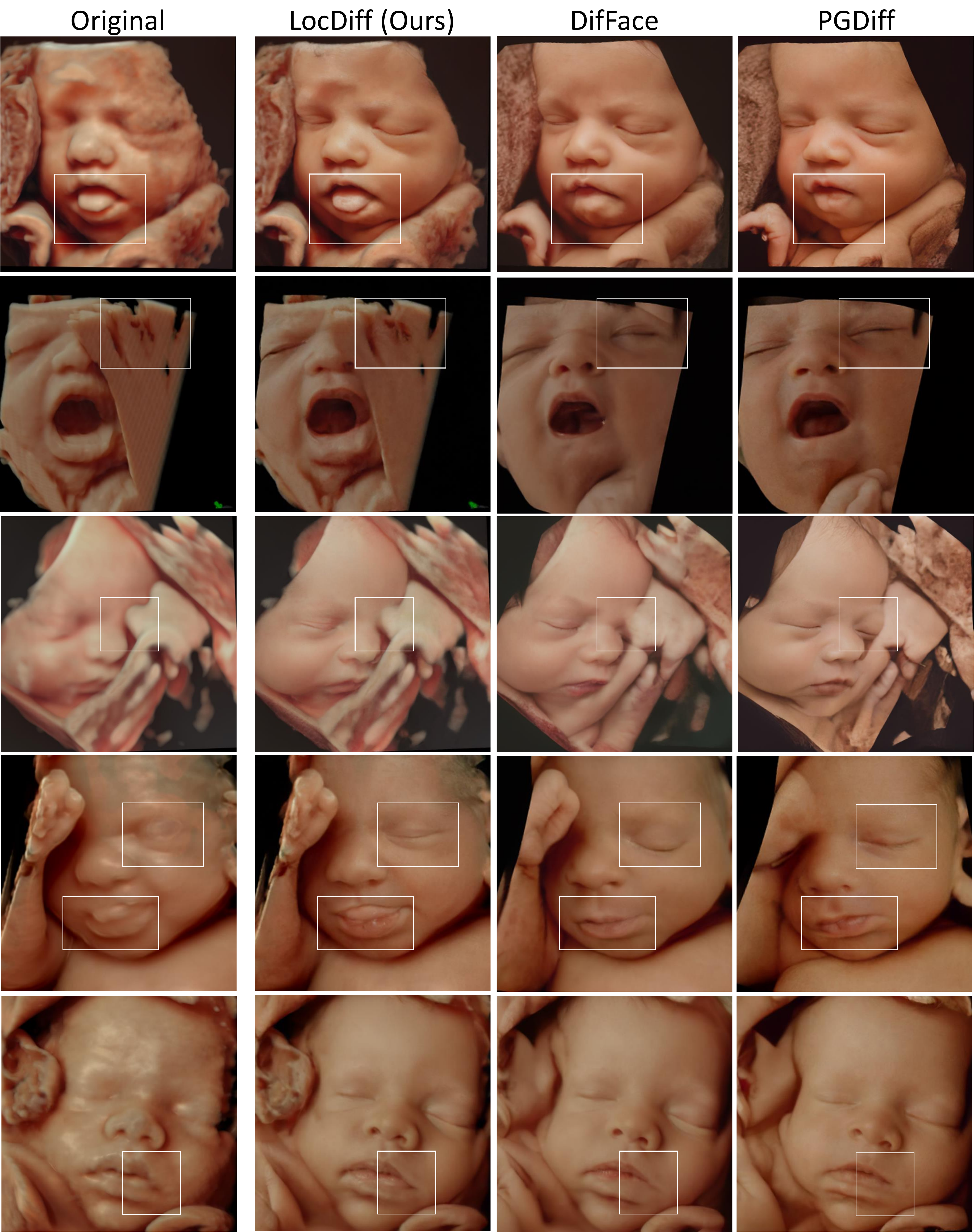}
	
	\caption{\textbf{Rendering-to-photo:} Qualitative comparison to baseline methods. LocDiff preserves facial characteristics better, as evidenced by the highlighted regions. Zoom in for best view.
	}

	\label{fig:bf_baselines}
	
\end{figure}

\subsection{Comparison with State-of-the-Art Methods} 
\label{sec:comparison_sota}
\noindent We compare our proposed method \textit{LocDiff} with SOTA methods for the art-to-photo and rendering-to-photo tasks. We employ DifFace~\cite{yueDifFaceBlindFace2024a} and PGDiff~\cite{yangPGDiffGuidingDiffusion2023} with the same model weights (FFHQ prior for art-to-image, fine-tuned baby face prior for rendering-to-photo) as LocDiff to facilitate a fair comparison and to evaluate the benefit of locality-constrained guidance over classifier guidance and guidance without flexible region constraints.   
For the art-to-photo task, we further compare our approach with recent blind face restoration methods, including DiffBIR~\cite{lin2024diffbir} and DT-BFR~\cite{kuai2025blindface}, as well as with models specifically designed for art painting enhancement, such as Art2Real~\cite{tomeiArt2RealUnfoldingReality2019} and ILVR~\cite{choiILVRConditioningMethod2021}. 
Additionally, we evaluate ControlNet\cite{zhangAddingConditionalControl2023}, a controllable text-to-image diffusion method with structure-preserving guidance using the tile-based control method. 

For SOTA implementations, we adopt the recommended settings and use the validation split for hyperparameter tuning (see \ref{sec:appendix_datasets}).

\textbf{Rendering-to-photo.} Table~\ref{tab:sota_bf} presents quantitative results on the ultrasound dataset, with visual comparisons in Fig.~\ref{fig:bf_baselines}. Quantitative and qualitative results demonstrate that our method achieves superior facial detail preservation while maintaining quality (FID) on par with baselines.

\begin{figure*}[h!]
	
	\centering

	\includegraphics[trim=0 0 0 0, clip,width=1\linewidth]{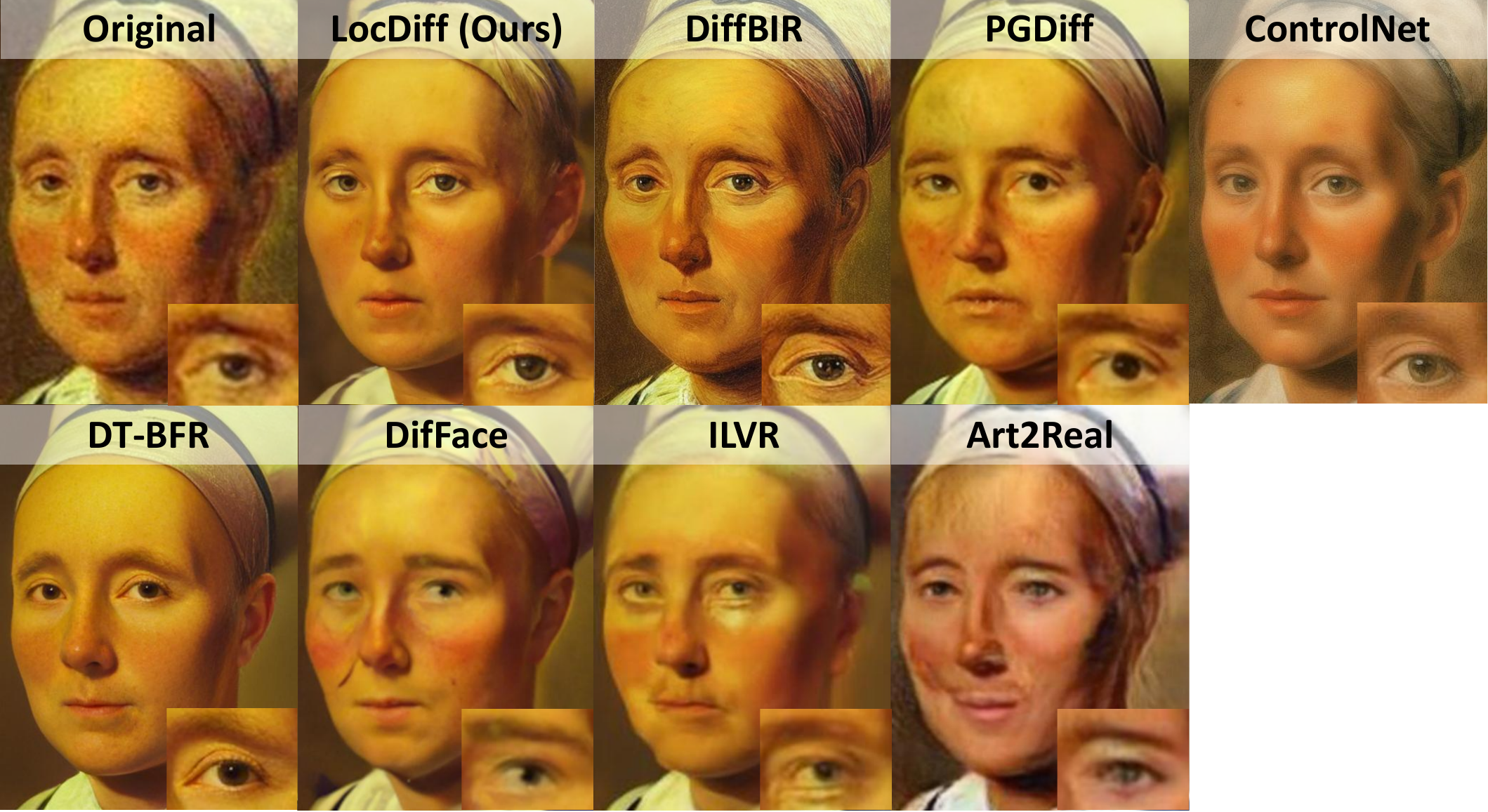}
	
	\caption{\textbf{Art-to-photo:} Qualitative comparison with state-of-the-art methods. Our proposed LocDiff enhances the low-quality art painting with photo-realistic details, while preserving facial identity.
	}
	\label{fig:sota_comparison_one_sample}
	
\end{figure*}

\begin{figure*}[h!]
	
	\centering

	\includegraphics[trim=0 0 0 0, clip,width=1\linewidth]{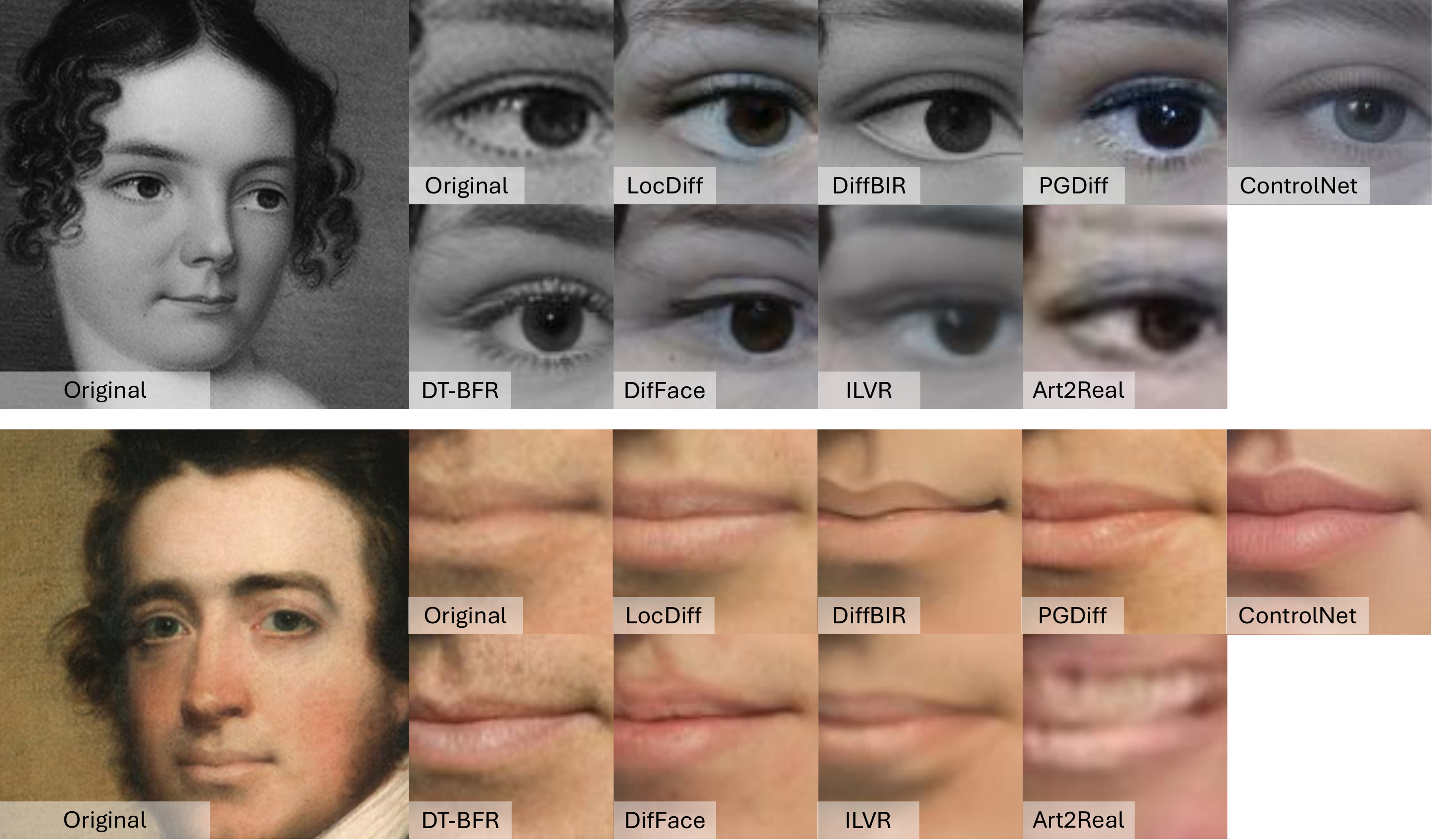}
	
	\caption{Visual comparison with baseline methods. Although DiffBIR excels at denoising and preserving structure, it lacks photorealistic details in the mouth (bottom) and eye (top) regions and retains a painting-like appearance. In contrast, our method (LocDiff) achieves photorealistic domain transfer while maintaining competitive identity preservation.
	}
	\label{fig:diffbir_comparison}
	
\end{figure*}

\textbf{Art-to-image.} Table \ref{tab:sota} presents quantitative comparisons on the \textit{MetFaces} and \textit{WikiArt-Faces} datasets. Figures~\ref{fig:sota_comparison_one_sample} and~\ref{fig:diffbir_comparison} show qualitative comparisons. 
While DifFace and PGDiff yield photorealistic outputs with favorable FID scores, they fail to preserve identity, as illustrated in Fig.~\ref{fig:sota_comparison_one_sample}. Conversely, DiffBIR~\cite{lin2024diffbir} achieves high identity-preserving metrics, demonstrating reliable performance in image denoising of faces. However, DiffBIR produces less photorealistic images, retaining fine brushstroke artifacts and thus exhibiting high FID, see Fig.~\ref{fig:diffbir_comparison} for further visual examples. 
While ControlNet adds photo-realistic details, it underperforms on identity-preservation compared to face-specific methods, reflecting the challenges of adapting text-to-image control methods to identity-preserving enhancement tasks. We observed that text-guided spatial control requires careful prompt engineering and may produce overly smooth outputs that lack the fine-grained detail preservation needed for face enhancement.
As the objective of \textit{LocDiff} is to preserve salient source domain features, it is not feasible to achieve complete alignment with the target domain, resulting in a slightly higher FID score.
Our \textit{LocDiff} model demonstrates a favourable trade-off of identity preservation and image quality, reflected by the second best metrics for LMD, SSIM-F, and IDS.

\begin{table}
\scriptsize
\centering

\begin{tabular}{lll|l|ll}
  & \multicolumn{2}{c|}{\textit{faithfulness}} 
  & \multicolumn{1}{c|}{\textit{realism}} 
  & \multicolumn{2}{c}{\textit{traditional metrics}} \\ \hline

\multicolumn{1}{l|}{\textit{Methods}} 
& \multicolumn{1}{r}{SSIM-F$\uparrow$} 
& \multicolumn{1}{r|}{IDS$\downarrow$} 
& \multicolumn{1}{c|}{FID$\downarrow$}  
& \multicolumn{1}{r}{LPIPS$\downarrow$} 
& \multicolumn{1}{r}{PSNR$\uparrow$} \\ \hline

\multicolumn{1}{l|}{DifFace~\cite{yueDifFaceBlindFace2024a}}      
& {\ul 0.73$\pm$0.06}                  
& {\ul 51.0$\pm$7.9}                   
& \textbf{74.8}                         
& {\ul 0.194$\pm$0.04}                  
& {\ul 28.27$\pm$1.2} \\

\multicolumn{1}{l|}{PGDiff~\cite{yangPGDiffGuidingDiffusion2023}} 
& 0.69$\pm$0.06                        
& 54.9$\pm$8.1                         
& 76.5                                  
& 0.240$\pm$0.05                        
& 26.61$\pm$1.1 \\

\multicolumn{1}{l|}{\textbf{LocDiff (Ours)}}                      
& \textbf{0.78$\pm$0.05}               
& \textbf{45.4$\pm$7.7}                
& {\ul 75.2}                            
& \textbf{0.188$\pm$0.04}               
& \textbf{28.75$\pm$1.2} \\ \hline

\end{tabular}

\caption{Quantitative comparison on the fetal ultrasound rendering dataset. Best results in \textbf{bold}; second-best {\ul underlined}.
}
\label{tab:sota_bf}
\end{table}

\begin{table*}[]
\centering
\scriptsize

\begin{tabular}{l|rrr|r|rr}

   & \multicolumn{3}{c|}{\textit{faithfulness}} 
   & \multicolumn{1}{c|}{\textit{realism}} 
   & \multicolumn{2}{c}{\textit{traditional metrics}} \\ \hline

\multicolumn{1}{l|}{\textit{Methods}} 
& LMD$\downarrow$ 
& SSIM-F$\uparrow$ 
& IDS$\downarrow$ 
& FID$\downarrow$  
& LPIPS$\downarrow$ 
& PSNR$\uparrow$ \\ \hline

\multicolumn{7}{c}{\textit{WikiArt-Faces dataset}} \\

Art2Real~\cite{tomeiArt2RealUnfoldingReality2019} 
& 6.6$\pm$8.1 & 0.50$\pm$0.1 & 51.5$\pm$11.3 & 130.4 
& {\ul 0.27$\pm$0.07} & 18.7$\pm$3.1 \\

ILVR~\cite{choiILVRConditioningMethod2021}        
& 4.5$\pm$5.7 & 0.56$\pm$0.1 & 46.0$\pm$7.1 & 108.6 
& \textbf{0.26$\pm$0.06} & {\ul 26.7$\pm$2.0} \\

DifFace~\cite{yueDifFaceBlindFace2024a}           
& 5.1$\pm$7.2 & 0.53$\pm$0.1 & 50.8$\pm$7.6 & \textbf{66.3} 
& 0.30$\pm$0.05 & 25.3$\pm$1.8 \\

PGDiff~\cite{yangPGDiffGuidingDiffusion2023}      
& 4.7$\pm$7.0 & 0.51$\pm$0.1 & 47.7$\pm$7.7 & {\ul 68.2} 
& 0.31$\pm$0.06 & 24.9$\pm$2.0 \\

DiffBIR~\cite{lin2024diffbir}                     
& \textbf{2.3$\pm$4.8} & {\ul 0.59$\pm$0.1} & \textbf{27.9$\pm$8.4} & 95.6 
& 0.39$\pm$0.10 & \textbf{28.0$\pm$2.8} \\

DT-BFR~\cite{kuai2025blindface}                   
& 4.6$\pm$7.1 & 0.48$\pm$0.1 & 47.1$\pm$7.6 & 74.3 
& 0.31$\pm$0.07 & 25.3$\pm$2.0 \\
 ControlNet~\cite{zhangAddingConditionalControl2023}& 5.1$\pm$7.1& 0.42$\pm$0.0& 52.2$\pm$7.7& 92.7& 0.40$\pm$0.08&22.7$\pm$2.0\\

\textbf{LocDiff (Ours)}                           
& {\ul3.9$\pm$7.3}& \textbf{0.61$\pm$0.1} & {\ul 43.5$\pm$7.2} & 70.9 
& 0.32$\pm$0.06 & 25.2$\pm$2.3 \\ \hline

\multicolumn{7}{c}{\textit{MetFaces dataset}} \\

Art2Real~\cite{tomeiArt2RealUnfoldingReality2019} 
& 5.1$\pm$3.5 & 0.45$\pm$0.2 & 49.4$\pm$15.2 & 114.5 
& 0.34$\pm$0.07 & 17.5$\pm$3.9 \\

ILVR~\cite{choiILVRConditioningMethod2021}        
& 4.4$\pm$1.8 & 0.53$\pm$0.1 & 46.7$\pm$7.9 & 74.5 
& 0.31$\pm$0.08 & {\ul 26.0$\pm$2.1} \\

DifFace~\cite{yueDifFaceBlindFace2024a}           
& 5.3$\pm$2.7 & 0.50$\pm$0.1 & 52.2$\pm$8.9 & \textbf{59.1} 
& 0.27$\pm$0.06 & 24.6$\pm$1.9 \\

PGDiff~\cite{yangPGDiffGuidingDiffusion2023}      
& 4.4$\pm$1.9 & 0.53$\pm$0.1 & 47.4$\pm$7.9 & 66.1 
& {\ul 0.26$\pm$0.05} & 24.3$\pm$2.1 \\

DiffBIR~\cite{lin2024diffbir}                     
& \textbf{1.6$\pm$1.0} & \textbf{0.62$\pm$0.1} & \textbf{18.3$\pm$5.3} & 99.3 
& {\ul 0.26$\pm$0.06} & \textbf{27.8$\pm$3.1} \\

DT-BFR~\cite{kuai2025blindface}                   
& 4.0$\pm$1.8 & 0.49$\pm$0.1 & 42.0$\pm$7.2 & 73.6 
& \textbf{0.22$\pm$0.04} & 25.0$\pm$2.2 \\
 ControlNet~\cite{zhangAddingConditionalControl2023}& 4.4$\pm$1.6& 0.42$\pm$0.1& 48.4$\pm$7.3& 97.9& 0.36$\pm$0.06&23.0$\pm$2.0\\

\textbf{LocDiff (Ours)}                           
& {\ul 3.2$\pm$1.6}& {\ul 0.60$\pm$0.1} & {\ul 41.0$\pm$7.6} & {\ul 64.8} 
& 0.27$\pm$0.06 & 24.6$\pm$2.3 \\ \hline

\end{tabular}

\caption{Quantitative comparison to SOTA methods on WikiArt-Faces and MetFaces. 
Best results in \textbf{bold}; second-best {\ul underlined}.}
\label{tab:sota}
\end{table*}

\textbf{Computational Requirements.} 
The locality-constrained guidance introduces a 1.4-1.7× overhead compared to baseline diffusion sampling (Tab. \ref{tab:computational_cost}), which we consider acceptable for offline enhancement tasks prioritizing quality over speed. Total inference time (7.96-16.84s per image) remains competitive with other gradient-based methods.

\begin{table}[h]
\scriptsize
\centering
\begin{tabular}{@{}lrr@{}}

Methods  & Time (s) & Peak GPU Memory (GB)  \\ 
  \hline
\multicolumn{3}{c}{\textit{Art-to-photo task (WikiArt/MetFaces)}} \\
Art2Real & 0.14 $\pm$ 0.01 & 8.3GB \\
ILVR & 0.64 $\pm$ 0.02 & 2.7GB \\
DifFace & 7.20 $\pm$ 0.17 & 5.8GB \\
PGDiff & 103.47 $\pm$ 14.28 & 4.8GB \\
DifBIR & 15.82 $\pm$ 0.59 & 13.9GB \\
DT-BFR & 0.17 $\pm$ 0.01 & 1.0GB \\
ControlNet & 3.41 $\pm$ 0.25 & 2.8GB \\
\noalign{\hrule height 0.1pt}
LocDiff (Ours) & 16.84 $\pm$ 0.28 & 6.2GB \\
Baseline diffusion (no LocDiff guidance) & 9.7 $\pm$ 0.19 & 5.8GB \\
\hline

\\[0.6em]

%\hline
\multicolumn{3}{c}{\textit{Rendering-to-photo task (Ultrasound)}} \\
DifFace & 7.29 $\pm$ 0.16 & 5.8GB \\
PGDiff & 101.74 $\pm$ 14.50 & 4.8GB \\
\noalign{\hrule height 0.1pt}
LocDiff (Ours) & 7.96 $\pm$ 0.18 & 8.2GB \\
Baseline diffusion (no LocDiff guidance) & 5.5 $\pm$ 0.15 & 5.8GB \\
\hline

\end{tabular}
\caption{Computational requirements of baseline methods. Time: average inference time per sample.}
\label{tab:computational_cost}
\end{table}

\textbf{Balancing Realism and Faithfulness.}
While the similarity metrics LPIPS and PSNR are useful for measuring overall image similarity, our primary interest lies in preserving facial identity. To this end, we are particularly concerned with the preservation of facial regions that are crucial for facial recognition. 
In the regions of eyes, nose, and mouth, even minor alterations can result in significant changes to the face, leading to unfaithful results. In other regions, such as hair or background, moderate differences may even be advantageous and result in a faithful outcome, see Fig.~\ref{fig:comparison_locality-constrained_vs_entire_image}. 
For instance, in the pursuit of a smooth skin effect, it may be preferable to prioritize a more seamless appearance over the preservation of minute structural details, such as brush strokes. 
Figure~\ref{fig:appendix_locality_constrained_face_parts} visualizes that the \textit{LocDiff} model ensures a high structural similarity for eyes, nose, and mouth, but provides less guidance strength for the background and hair region. This results in a smoother background and high-frequency details for hairs, thereby adapting to the photorealistic domain. 
As previously stated, the realism-faithfulness trade-off in the context of domain shifts implies that the more realistic the results become in the target domain, the less we can preserve the source domain. To achieve faithful \textit{and} realistic image enhancement, it is essential to achieve a balance between these two. 
The results demonstrate that our proposed locality-constraint guidance 
demonstrates a favourable trade-off of realism and identity preservation, adapting to user preferences and improving image enhancement under domain shifts.

\subsection{Robustness to Segmentation Mask Quality} 
\label{sec:ablation_segmentation_mask}
\noindent We conduct a comprehensive ablation study examining both the performance of state-of-the-art face parsing models on our domain-shifted datasets and the sensitivity of LocDiff to mask inaccuracies.
\\
\noindent \textbf{Face Parsing Model Performance on Domain-Shifted Data.}
We evaluate three state-of-the-art face parsing models: FaRL~\cite{zheng2022farl} trained on LaPa~\cite{liu2020new}, FaRL trained on CelebAMask-HQ~\cite{karras2018progressive}, and SegFace~\cite{narayan2025segface} trained on LaPa. We extract six semantic classes (eyes, eyebrows, nose, mouth, skin) following our standard preprocessing pipeline (Section~\ref{experiments_regions}), with remaining classes treated as background.

\begin{table}[h!]
\centering
\scriptsize
\begin{tabular}{@{}llccccc@{}}
\toprule
\textbf{Dataset} & \textbf{Model} & R.Eye & L.Eye & Nose & Mouth & Brows \\
\midrule
\multirow{3}{*}{\textbf{MetFaces}} 
& SegFace (LaPa) & 0 & 0 & 0 & 0 & 0 \\
& FaRL (LaPa) & 0 & 0 & 0 & 0 & 0 \\
& FaRL (CelebA-HQ) & 14 & 8 & 0 & 0 & 0 \\
\midrule
\multirow{3}{*}{\textbf{WikiArt}} 
& SegFace (LaPa) & 0 & 1 & 0 & 0 & 0 \\
& FaRL (LaPa) & 0 & 0 & 0 & 0 & 0 \\
& FaRL (CelebA-HQ) & 10 & 9 & 0 & 0 & 0 \\
\midrule
\multirow{3}{*}{\textbf{Ultrasound}} 
& SegFace (LaPa) & 8 & 13 & 0 & 0 & 1 \\
& FaRL (LaPa) & 1 & 3 & 0 & 2 & 1 \\
& FaRL (CelebA-HQ) & 330 & 339 & 0 & 2 & 10 \\
\bottomrule
\end{tabular}

\vspace{0.3cm}

\begin{tabular}{@{}lcccc@{}}
\toprule
\textbf{Dataset} & FaRL (L) vs. & FaRL (L) vs. & FaRL (C) vs. & Avg \\
 & FaRL (C) & SegFace (L) & SegFace (L) & mIoU \\
\midrule
MetFaces & 0.818 & 0.876 & 0.854 & 0.849 \\
WikiArt & 0.846 & 0.883 & 0.869 & 0.866 \\
Ultrasound & 0.473 & 0.574 & 0.518 & 0.522 \\
Ultrasound (w/o eyes) & 0.873 & 0.848 & 0.818 & 0.846 \\
\bottomrule
\end{tabular}
\caption{\textbf{Top:} Missing semantic classes across face parsing models. \textbf{Bottom:} Pairwise cross-model agreement (mIoU). L=LaPa, C=CelebAMask-HQ. Dataset sizes: MetFaces (1366), WikiArt (1373), Ultrasound (437). High mIoU indicates reliable parsing despite domain shift.}
\label{tab:missing_features}
\end{table}

Table~\ref{tab:missing_features} reports the frequency of missing semantic classes across models and datasets. On MetFaces and WikiArt, all LaPa-trained models achieve highly accurate detection, while the CelebAMask-HQ model struggles with closed eyes (14/8 missing eye detections on MetFaces). This may be explained by the fact that closed eyes are underrepresented in the CelebAMask-HQ dataset. The ultrasound dataset is the most challenging due to the fact that most fetuses have their eyes closed. However, FaRL (LaPa) maintains robust performance with only 1/3 missing eye detections compared to substantial failures of the CelebAMask-HQ model.

To quantify parsing consistency, we measure pairwise mean IoU (mIoU) between models across semantic classes (Tab.~\ref{tab:missing_features}, bottom). On MetFaces and WikiArt, models show strong agreement (average mIoU: 0.849 and 0.866), indicating reliable parsing on artistic images. On ultrasound data, overall agreement drops (average mIoU: 0.522) due to eye detection failures. However, when excluding eye classes, agreement remains high (average mIoU: 0.846), demonstrating that parsing quality for other facial regions is maintained despite significant domain shift.

FaRL (LaPa) demonstrates the best generalization across all datasets and semantic classes, which is also used for all other experiments in this paper. 
We illustrate how parsing variations/failures affect enhancement quality on samples from the MetFaces and Ultrasound datasets with low cross-model mIOU to, see Figs.~\ref{fig:parsing_model_ablation_metfaces} and~\ref{fig:parsing_model_ablation_ultrasound}.

\noindent \textbf{Sensitivity to Mask Perturbations.}
To systematically analyze robustness to mask inaccuracies, we evaluate LocDiff under controlled mask degradation scenarios on MetFaces and Ultrasound datasets:
\begin{itemize}
    \item \textit{Spatial misalignment}: 10-pixel shifts in x/y directions
    \item \textit{Noise}: 30\% random pixels reassigned to background
    \item \textit{Morphological errors}: Dilation (10 pixels), erosion (5/10 pixels)
    \item \textit{Missing features}: Eye classes assigned to skin or background class
\end{itemize}

\begin{table}[h!] 
\centering
\scriptsize
\begin{tabular}{@{}lcccc@{}}
\midrule
\textbf{Mask Perturbation} & LMD $\downarrow$ & SSIM-F $\uparrow$ & IDS $\downarrow$ & FID $\downarrow$ \\
\midrule
\multicolumn{5}{c}{\textit{MetFaces dataset}} \\
None & $3.2 \pm 1.6$ & $0.60 \pm 0.11$ & $41.0 \pm 7.6$ & $64.8$ \\
XY-shift (10 pixels) & $3.3 \pm 1.5$ & $0.59 \pm 0.10$ & $41.5 \pm 7.6$ & $65.4$ \\
Noise (30\%) & $3.6 \pm 1.6$ & $0.57 \pm 0.10$ & $42.8 \pm 7.8$ & $63.6$ \\
Dilation (10 pixels) & $3.0 \pm 1.4$ & $0.59 \pm 0.10$ & $38.2 \pm 7.3$ & $66.6$ \\
Erosion (5 pixels) & $4.1 \pm 1.7$ & $0.54 \pm 0.10$ & $46.9 \pm 8.0$ & $63.6$ \\
Eyes to background & $4.3 \pm 1.8$ & $0.53 \pm 0.11$ & $45.0 \pm 7.8$ & $64.0$ \\
Eyes to skin & $3.7 \pm 1.6$ & $0.56 \pm 0.10$ & $43.1 \pm 7.8$ & $63.8$ \\
\midrule
\multicolumn{5}{c}{\textit{Ultrasound Rendering dataset}} \\
None & -- & $0.78 \pm 0.05$ & $45.4 \pm 7.7$ & $75.2$ \\
XY-shift (10 pixels) & -- & $0.77 \pm 0.06$ & $44.3 \pm 8.0$ & $74.6$ \\
Noise (30\%) & -- & $0.79 \pm 0.05$ & $41.8 \pm 7.6$ & $75.3$ \\
Dilation (10 pixels) & -- & $0.75 \pm 0.06$ & $46.4 \pm 8.2$ & $73.3$ \\
Erosion (10 pixels) & -- & $0.77 \pm 0.06$ & $44.2 \pm 8.1$ & $75.6$ \\
Eyes to background & -- & $0.81 \pm 0.05$ & $40.7 \pm 7.2$ & $76.3$ \\
Eyes to skin & -- & $0.77 \pm 0.06$ & $44.4 \pm 8.0$ & $74.6$ \\
\midrule
\end{tabular}
\caption{Sensitivity to mask perturbations on MetFaces and Ultrasound datasets. Metrics (mean $\pm$ std) show LocDiff is robust to common parsing errors, with performance degrading less than 10\% in most cases. Metric changes reflect how perturbations redistribute pixels between regions with different conditioning strengths (MetFaces: strong eyes/skin, weak background; Ultrasound: strong background, weak eyes). Perturbations reducing strongly-conditioned regions trade faithfulness for photorealism, while the reverse occurs when strongly-conditioned regions expand. Visual examples in Figs.~\ref{fig:ablation_mask_degradation_1} and \ref{fig:ablation_mask_degradation_2}.
}
\label{tab:mask_sensitivity_metfaces}
\end{table}

Table~\ref{tab:mask_sensitivity_metfaces} and Figs.~\ref{fig:ablation_mask_degradation_1} and~\ref{fig:ablation_mask_degradation_2} present quantitative and qualitative results. LocDiff maintains robust performance under spatial misalignment, moderate noise, and small morphological perturbations, with metrics degrading by less than 10\% in most cases. The most significant degradation occurs when eyes are entirely missing from the face parsing map, though enhancement quality remains acceptable as shown in the difference maps. These results demonstrate that LocDiff is resilient to common parsing errors encountered in practice and moderate inaccuracies of face parsing maps do not lead to artifacts.

\subsection{Limitations}
\label{sec:limitations}
The degree of realism depends on the power of the pre-trained diffusion model. Consequently, strong domain shifts between training and source images will result in an unrealistic appearance. 
From our observations, performance quality is limited for unnatural face shapes, as seen in surrealistic paintings or statues, as well as images characterized by a high degree of noise or distortion, see Fig.~\ref{fig:failure_cases1}.

\section{Conclusion and Future Work}
\label{sec:conclusion}

Motivated by the challenge of balancing realism and faithfulness in cross-domain image enhancement, we presented a locality-constrained guidance approach named \textit{LocDiff}. Our method enables region-adaptive control of pre-trained diffusion models, offering flexible integration that supports both fully zero-shot scenarios and scenarios with lightweight prior alignment.

To enable diffusion models trained on photos to handle input images with domain shifts, such as art paintings or renderings, we extend the sampling step of the reversed diffusion process with locality-constrained guidance. This is crucial to preserve selected local source-domain features while enabling global adaptation of the remaining features to the target domain.
Our method enables precise control over the image enhancement process and facilitates user adaptation, e.g., adjusting of the local image region by modifying the attribution map and adapting the guidance strength.

Our experimental results demonstrate that \textit{LocDiff} performs competitively with state-of-the-art methods, achieving effective realism-faithfulness trade-offs on both art painting and ultrasound rendering datasets, addressing the distinct challenge of cross-domain enhancement rather than in-domain restoration.

\section*{Declaration of generative AI and AI-assisted technologies in the manuscript preparation process}
\noindent During the preparation of this work the authors used DeepL to refine the language and improve readability. After using this tool, the author reviewed and edited the content as needed and take full responsibility for the content of the published article.

%% Use \subsubsection, \paragraph, \subparagraph commands to 
%% start 3rd, 4th and 5th level sections.
%% Refer following link for more details.
%% https://en.wikibooks.org/wiki/LaTeX/Document_Structure#Sectioning_commands


\section*{Acknowledgements}
\noindent The VRVis GmbH is funded by BMIMI, BMWET, Tyrol, Vorarlberg and Vienna Business Agency in the scope of COMET - Competence Centers for Excellent Technologies (911654) which is managed by FFG.

\appendix

\section{Appendix}
\label{sec:appendix}
\subsection{Datasets}
\label{sec:appendix_datasets}
All images in the used datasets are aligned
to the FFHQ~\cite{karrasStyleBasedGeneratorArchitecture2019a} template face and resized to $512\times512$ pixels using an alignment method\footnote{github.com/xinntao/facexlib}.
We used the provided code\footnote{huggingface.co/datasets/asahi417/wikiart-face} to curate the \textit{WikiArt-Faces} dataset. 

\noindent \textit{Baby Portrait Dataset (100 images):}
In order to fine-tune the diffusion prior for the rendering-to-photo task, we collected a small dataset comprising 100 high-resolution face portrait images of babies. 
The images were sourced from \href{Pexels.com}{Pexels.com} using the search terms “baby” and “newborn.” Selection was restricted to images under a CC0 license with a single clearly visible face, only minimal facial occlusion  (no objects or hairs covering important facial features), a balanced representation of head poses with 50\% frontal and 50\% non-frontal views, and high image quality. Images matching the search terms were manually reviewed, and the first 100 that satisfied the selection criteria were downloaded. All images were then preprocessed using the same alignment protocol described above.

\subsection{Hyperparameter \& Baseline Implementation}
For each dataset, we randomly selected 20\% of the data to perform hyperparameter search. The performance metrics for the ablation study and the baselines were subsequently reported on the remaining 80\% of the data.
We evaluated the metrics of ControlNet~\cite{zhangAddingConditionalControl2023} with the tile-based control method combined with the Realistic Vision 5.1 backbone, using the positive prompt "best quality, highly detailed face, photorealistic portrait" and negative prompt "blurry, low quality, distorted"), control scale 1.0, and guidance scale 5.
We tested multiple configurations including various control types (tile, canny, soft boundaries) and backbones (SD 1.5, Realistic Vision 5.1) and report results for the best setting above.

\subsection{Metrics}
\label{sec:appendix_metrics}
IDS leverages ArcFace~\cite{dengArcFaceAdditiveAngular2019} embeddings to measure the angular distance between the features of the ground truth and reference images. The implementation from \cite{wangRestoreFormerRealWorldBlind2023a} is employed for this computation.
To compute the Landmark Distance (LMD) score, which quantifies the distance between facial landmarks, we use SPIGA~\cite{prados-torreblancaShapePreservingFacial2022} to predict the facial landmarks and choose only the landmarks for eyebrows, eyes, nose, and mouth, resulting in a total of 65 landmarks.
We use the torchmetrics framework\footnote{github.com/Lightning-AI/torchmetrics} to calculate LPIPS, PSNR, and SSIM. To calculate the FID distance we use the implementation of \cite{stein2023exposing}. We compute the SSIM-F as the mean SSIM over the eye, nose, and mouth regions. Peak GPU memory during inference (batch size = 1, 512×512 input) was measured on an Nvidia A100 using \textit{torch.cuda.max\_memory\_allocated}.

\subsection{Implementation Details: Art-to-photo}
\label{sec:appendix-art2photo}

For the art-to-photo task, Fig. \ref{fig:x0_sampling} illustrates that at time step $t=60$, the predicted image $x_{0|60}$ exhibits comparable image semantics to the final predicted image $x_{0|0}$. We terminate the guidance at step 60 for our experiments, because the remaining steps only have minimal impact on the image semantics, as they are designed to develop high-frequency details.

\begin{itemize}
    \item Eye, eyebrow, mouth, nose regions: $(t_{min} = 60, m = 2)$
    \item Skin region: $(t_{min} = 60, m = 4)$
    \item Background region: $(t_{min} = 90, m = 8)$
\end{itemize}
We found that metrics remain stable beyond $140$ steps, see Tab.~\ref{tab:ablation_hyperparameter}, so we set the starting time step $T_{start}=140$. \\ 

\subsection{Implementation Details: Rendering-to-photo}
\label{sec:appendix-rendering2photo}

For the rendering-to-photo task, the goal is to preserve the ultrasound appearance while incorporating photorealistic details with moderate enhancement. This contrasts with the art-to-photo task, which aims for a result closely aligned with the photorealistic domain. To achieve this, we set $T_{start}=80$ and stop guidance at $t_{min}=40$.
In the ultrasound domain, eye details are often not visible. Therefore, we apply a weak guidance strength ($m=8$) for the eyes to avoid compromising the generation of high-resolution details. For these barely visible features, the model is encouraged to generate plausible eye details. We apply strong constraints to the background, ensuring it closely matches the input image to prevent the generation of artifacts in the surrounding tissue.
\begin{itemize}
    \item Eye and skin regions: $(t_{min} = 40, m = 8)$
    \item Mouth and nose regions: $(t_{min} = 40, m = 4)$
    \item Background region: $(t_{min} = 40, m = 2)$.
\end{itemize}

\subsection{Additional Visual Results}
\begin{figure*}[h!]
\centering
\includegraphics[trim=0 0 0 0, clip, width=1\linewidth]{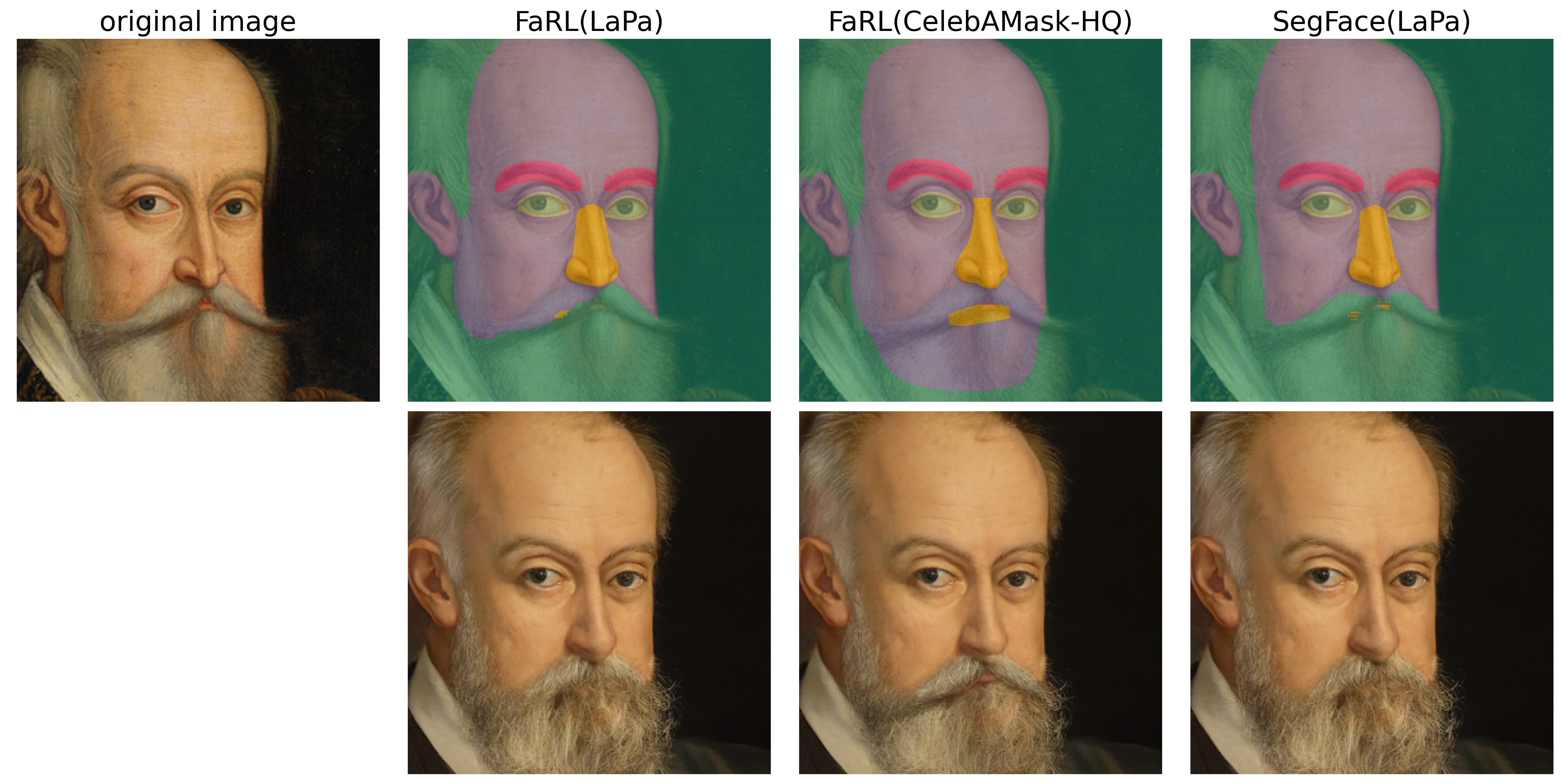}
\caption{\textbf{Impact of parsing disagreement on challenging case.} MetFaces sample with lowest cross-model mIoU (0.709) in the test set. Parsing disagreement around the mouth region (mouth vs. skin vs. background) produces visible differences in enhancement: stronger conditioning on mouth/skin regions preserves the moustache better than weak background conditioning. Despite incorrect parsing, no artifacts are introduced in the enhanced images.}
\label{fig:parsing_model_ablation_metfaces}
\end{figure*}

\begin{figure*}[h!]
\centering
\includegraphics[trim=0 0 0 0, clip, width=1\linewidth]{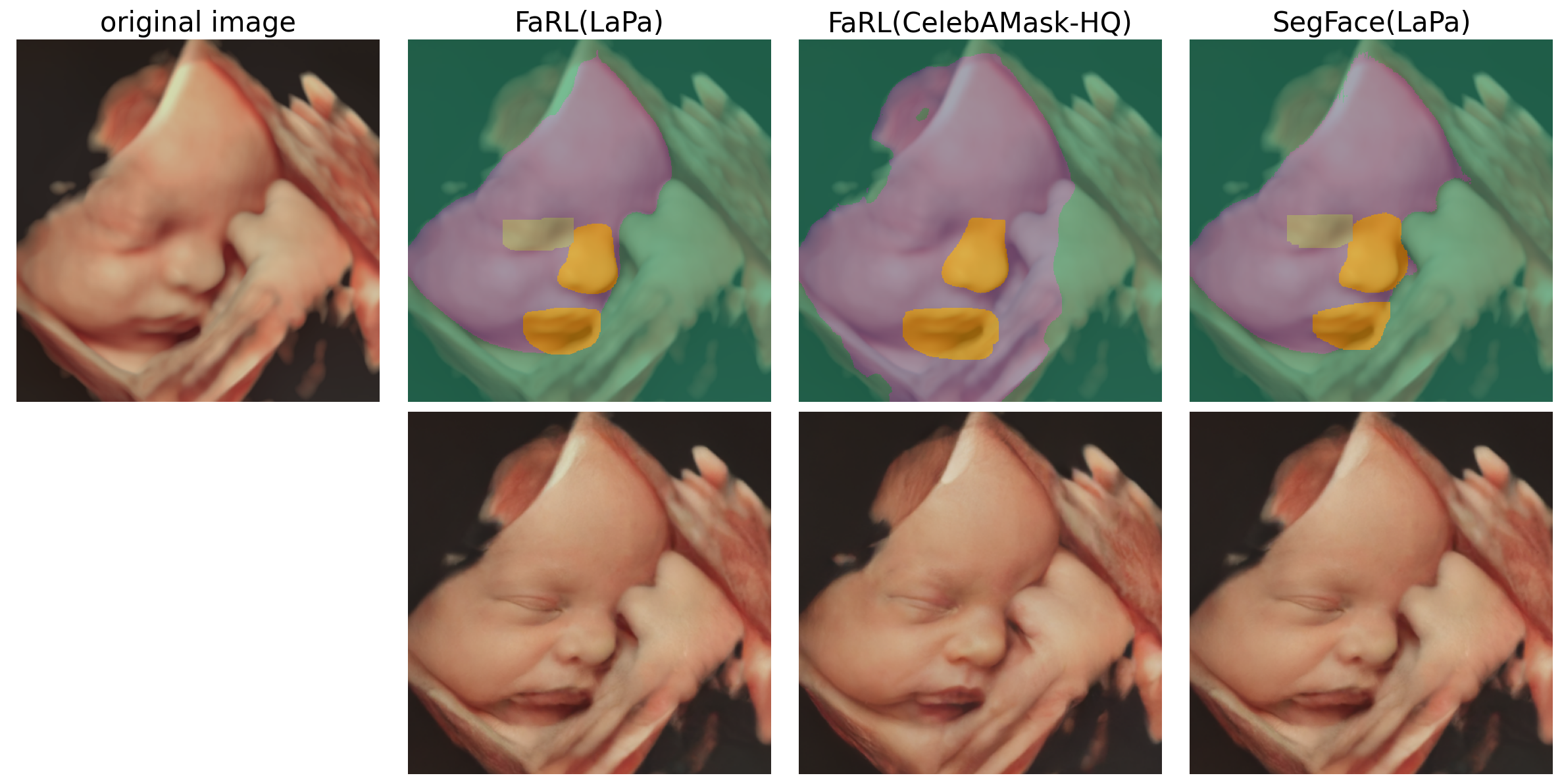}
\caption{\textbf{Impact of parsing disagreement on challenging ultrasound sample.} Ultrasound rendering with lowest cross-model mIoU (0.424) in the test set. FaRL (CelebM-HQ) incorrectly assigns eye and background regions to skin. Given the ultrasound-specific conditioning (strong background, weak skin/eyes), misclassified background regions are slightly more enhanced and the eye region is enhanced differently. Despite these mask errors, enhancement quality degrades moderately with only localized differences.}
\label{fig:parsing_model_ablation_ultrasound}
\end{figure*}

\begin{figure*}[h!]
\centering
\includegraphics[trim=0 0 0 0, clip, width=1\linewidth]{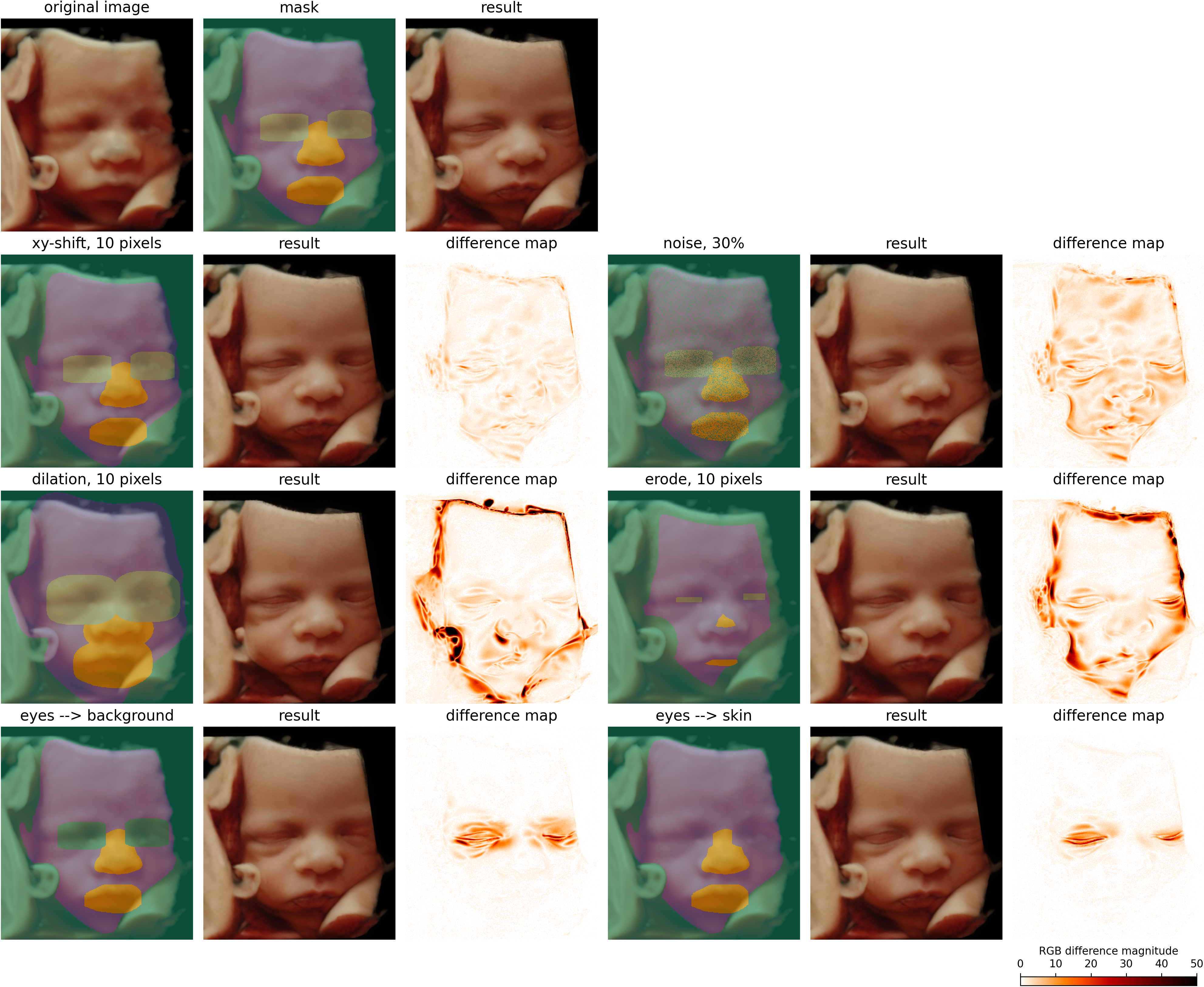}
\caption{\textbf{Mask perturbation robustness analysis on ultrasound data.} 
Top row: input rendering, unperturbed mask, and baseline enhancement. Rows 2-4 show mask perturbation pairs: perturbed mask, resulting enhancement, and RGB difference magnitude from baseline enhancement. Differences are most visible in regions where perturbations alter conditioning strength. For the ultrasound task, background receives strong conditioning while skin/eyes receive weak conditioning. Difference maps reveal that most perturbations cause subtle, localized changes only, with missing eye classes and severe morphological errors producing the most visible deviations.}
\label{fig:ablation_mask_degradation_1}
\end{figure*}

\begin{figure*}[h!]
\centering
\includegraphics[trim=0 0 0 0, clip, width=1\linewidth]{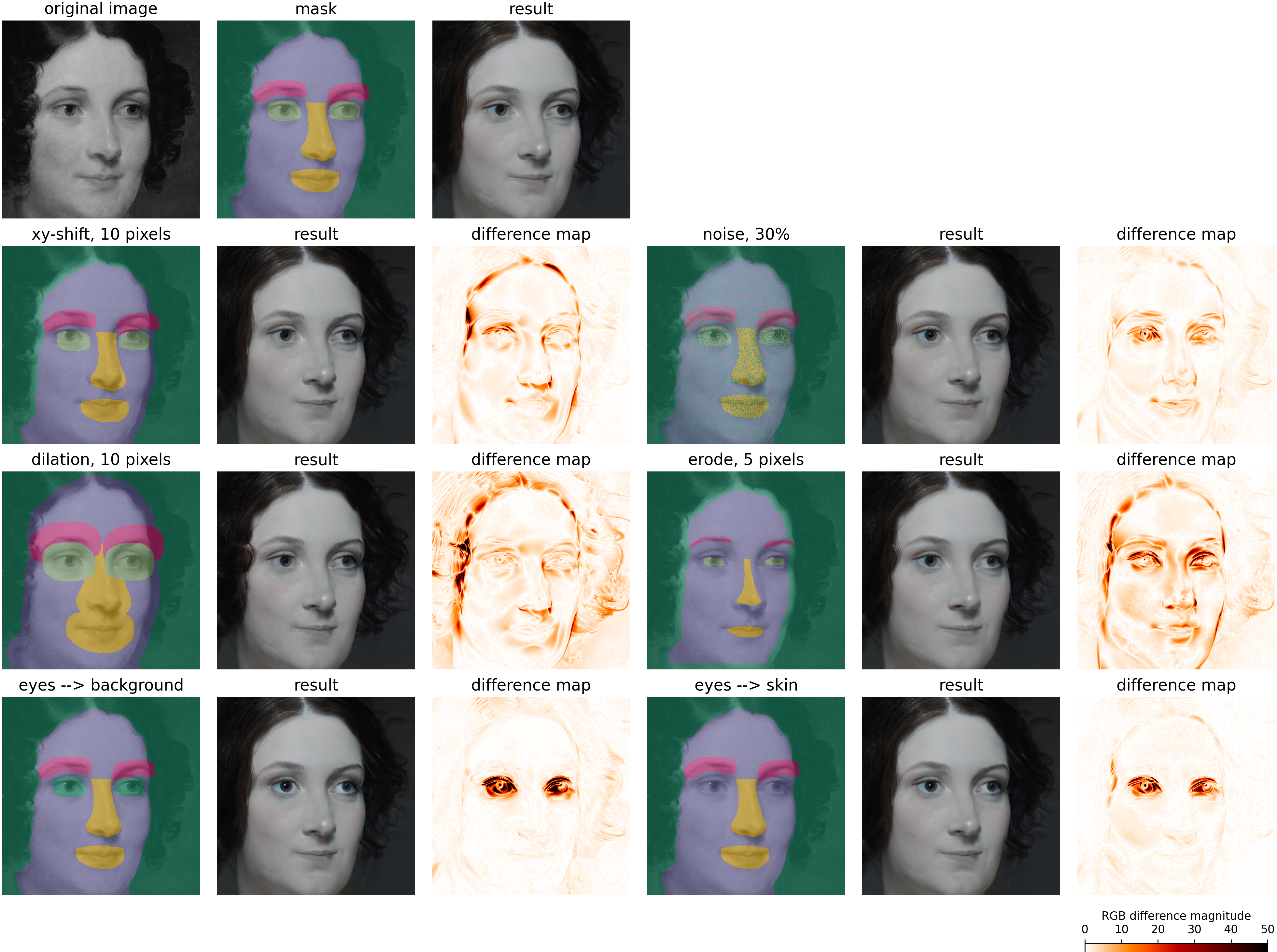}
\caption{ \textbf{Mask perturbation robustness analysis on MetFaces.} Top row: input, baseline mask, and baseline enhancement. Subsequent rows show pairs of perturbed masks, corresponding enhancements, and difference maps (RGB magnitude from baseline). Strongest differences occur in perturbed regions where conditioning strength changes, particularly when critical facial features like eyes are misclassified, though overall enhancement remains stable. The dilation perturbation better preserves hair structure, as the dilated skin class (more strongly conditioned than the background) partially covers the hair regions.}
\label{fig:ablation_mask_degradation_2}
\end{figure*}

\begin{figure*}[h!]
\centering
\includegraphics[trim=0 0 0 0, clip, width=1\linewidth]{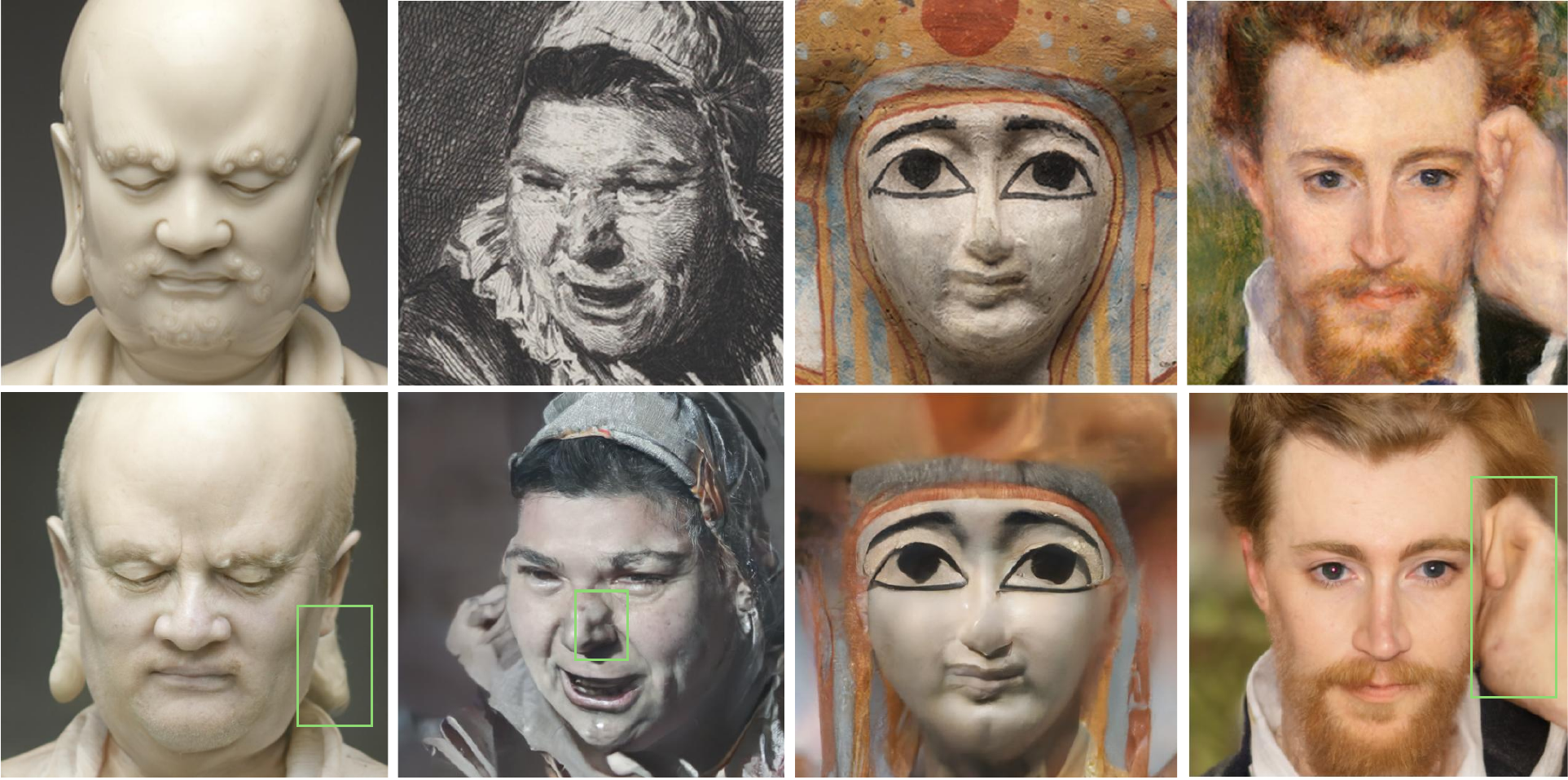}
\caption{\textbf{Failure cases of LocDiff}. Top row: original images. Bottom row: enhanced results. The method struggles when: (1) extreme anatomical deviations (elongated ear) are normalized to common shapes present in the training data, (2) heavy artistic brush strokes cause structural misinterpretation (nose deformation), (3) highly abstract face semantics produce only denoising without photorealistic detail generation, and (4) out-of-distribution elements like hands are distorted due to absence from the FFHQ training prior.}
\label{fig:failure_cases1}
\end{figure*}

Figures~\ref{fig:parsing_model_ablation_metfaces}, \ref{fig:parsing_model_ablation_ultrasound}, \ref{fig:ablation_mask_degradation_1}, \ref{fig:ablation_mask_degradation_2} support the mask quality 
ablation study presented in Section~\ref{sec:ablation_segmentation_mask}. 
Figure~\ref{fig:failure_cases1} presents failure cases corresponding to the limitations analyzed in Section~\ref{sec:limitations}.

\bibliographystyle{elsarticle-num} 
\bibliography{main}



\end{document}